\documentclass{article}

\usepackage[T1]{fontenc}
\usepackage[utf8]{inputenc}
\usepackage{indentfirst}
\usepackage{newpxtext}
\usepackage{amsmath}
\usepackage{amssymb}
\usepackage{amsthm}
\usepackage{graphicx}
\usepackage{bm}
\usepackage{mathtools}
\usepackage[left = 20 truemm , right = 20 truemm, bottom = 20 truemm]{geometry}
\usepackage{url}
\usepackage{algorithmic}
\usepackage{algorithm}
\usepackage{dcolumn}
\usepackage[hidelinks]{hyperref}
\usepackage{orcidlink}

\makeatletter
\let\originalcaption\caption
\def\captiontypealgorithm{algorithm}
\renewcommand{\caption}[2][]{%
    \def\captiontext{#2}%
    \ifx\@captype\captiontypealgorithm
        \def\captiontext{#2}%
    \else
        \def\captiontext{\textbf{#2}}%
    \fi
    \if\relax\detokenize{#1}\relax
        \originalcaption{\captiontext}%
    \else
        \originalcaption[#1]{\captiontext}%
    \fi
}
\renewcommand{\fnum@figure}{\textbf{Figure\nobreakspace\thefigure.}}
\renewcommand{\fnum@table}{\textbf{Table\nobreakspace\thetable.}}
\long\def\@makecaption#1#2{%
    \vskip\abovecaptionskip
    \sbox\@tempboxa{#1 #2}%
    \ifdim \wd\@tempboxa >\hsize
        #1 #2\par
    \else
        \global \@minipagefalse
        \hb@xt@\hsize{\hfil\box\@tempboxa\hfil}%
    \fi
    \vskip\belowcaptionskip
}
\makeatother

\newenvironment{affiliations}{%
    \setcounter{enumi}{1}%
    \setlength{\parindent}{0in}%
    \slshape\sloppy%
    \begin{list}{\upshape$^{\arabic{enumi}}$}{%
        \usecounter{enumi}%
        \setlength{\leftmargin}{0in}%
        \setlength{\topsep}{0in}%
        \setlength{\labelsep}{0in}%
        \setlength{\labelwidth}{0in}%
        \setlength{\listparindent}{0in}%
        \setlength{\itemsep}{0ex}%
        \setlength{\parsep}{0in}%
        }
    }{\end{list}\par\vspace{12pt}}

\title{
    Optimization of time-consuming experimental conditions using pseudo-experimental data guided by adaptive polynomial regression
}

\author{
Hirotaka Sugawara$^{1}$ \orcidlink{0009-0001-6823-1572},
Yujin Taguchi$^{1}$ \orcidlink{0009-0007-7997-6880},
Kei Minagawa$^{1}$,\\
Yusuke Hiki$^{1,2}$ \orcidlink{0000-0001-6955-3867},
Takashi Morikura$^{1}$ \orcidlink{0000-0003-1287-8809},
Akira Funahashi$^{1,2,*}$ \orcidlink{0000-0003-0605-239X}
}

\date{}

\begin{document}
\maketitle

\begin{affiliations}
    \item[$^{1}$] Graduate School of Science and Technology, Keio University, Kanagawa, Japan
    \item[$^{2}$] Department of Biosciences and Informatics, Keio University, Kanagawa, Japan
    \item[$^{*}$] funa@bio.keio.ac.jp
\end{affiliations}

\section*{Abstract}
Bayesian optimization (BO) is an optimization method that sequentially proposes the next candidate explainable variables for optimizing target variables by balancing exploration and exploitation.
BO is often used under a limited evaluation budget, such as hyperparameter tuning of deep learning.
Despite its effectiveness, conventional BO may have poor convergence in practical experimental science where each evaluation is often costly and time-consuming.
Recently, BO methods have been proposed that accelerate optimization by using pseudo-experimental data that simulate experimental data.
However, when only a limited number of experimental data are available, the generated pseudo-experimental data may be of insufficient quality.
In this study, we developed PolyBO to improve optimization time by generating high-quality pseudo-experimental data even when the number of trials is limited.
PolyBO performs BO efficiently by generating pseudo-experimental data with an adaptively updated versatile parametric model.
This low-capacity polynomial regression model is intended to enable efficient BO even with limited experimental data.
PolyBO updates the BO surrogate model with a combined dataset consisting of experimental data and pseudo-experimental data and then performs optimization.
Using synthetic benchmark functions with diverse landscapes, we found that PolyBO reduced the optimization time by a median of 42\%.
For a real-world material composition optimization problem, PolyBO reduced the optimization time by a median of 96\% compared with conventional methods.
Overall, PolyBO achieves efficient optimization in settings where each experiment requires a long time.

\clearpage

\section*{Introduction}
Optimization of experimental conditions, including material composition, has contributed to the discovery and design of new materials \cite{wang2023_CloseForm,Lookman2019_review,Tom2024_SDL}.
In particular, computational optimization of the development of biomaterials, which usually requires time-consuming experiments, has accelerated biological insights and applications across various biological fields, ranging from protein engineering \cite{arnold2017_DirectEvo} to cellular agriculture \cite{Quek2024_culturedMeat} and biopharmaceutical manufacturing \cite{Walsh2022_Biopharm,Hashizume2023_OptCulture}.
In the optimization of experimental conditions, the goal is to search for those that maximize the value of the objective function within a search space defined by the explainable variables of the target system.

Response surface methodology is a traditional design-of-experiments optimization method that approximates the objective function with a polynomial function and searches for optimal experimental conditions by sampling uniformly across the search space \cite{Bezerra2008_RSM}.
For example, using this approach, Skrivergaard et al. \cite{Skrivergaard2023_RSM} have optimized the concentrations of three serum-free medium components, fibroblast growth factor 2, fetuin, and bovine serum albumin, and formulated a medium that exceeds the cell-proliferation performance of a conventional medium containing 10\% fetal bovine serum.
However, (i) accurate modeling of target-system behavior is difficult because nonlinear interactions among explainable variables are intertwined \cite{wang2023_CloseForm,Cosenza2022_Multi-informationBO,Hashizume2024_ChallengesMLMedia}, and (ii) the relationship between explainable variables and objective values is a black-box relationship, and experiments for evaluating it are time- and resource-intensive \cite{Ndahiro2025_CHOHighCost,zheng2023_BioPharmHighCost}.
To handle complex nonlinear responses and efficiently identify promising experimental conditions with a limited number of experiments, an alternative method is needed.

In biology, Bayesian optimization (BO) \cite{kushner1964,garnett2023_BObook} meets these requirements \cite{Yang2019_arnoldBO,Hu2022_ProteinEngBO,gisperg2025_BioProcessBO}.
In BO, a surrogate model approximates the objective function, and selection of the next experimental condition is based on its predictive mean and uncertainty.
Because BO balances exploration of unobserved conditions and exploitation of past experimental results, it efficiently searches for and proposes experimental conditions \cite{Wang2023_RecentAdvBO}.
However, vanilla BO faces two challenges when applied to time-consuming experiments such as biological experiments.
First, often $>10$ explainable variables are involved \cite{Quek2024_culturedMeat,Hashizume2023_OptCulture,Cosenza2022_Multi-informationBO,Hashizume2024_ChallengesMLMedia,Narayanan2025}, and BO is particularly inefficient in such high-dimensional spaces \cite{Wang2023_RecentAdvBO,kandasamy2015_HDBOfirst,Tripp2020_W-LBO}.
Latent-space Bayesian optimization (LSBO) is a major strategy for addressing this challenge \cite{Tripp2020_W-LBO,Gomez-Bombarelli2018_LSBO,grosnit2021_T-LBO}.
LSBO performs BO in a $d$-dimensional latent space, which is obtained by strongly reducing the dimensionality of a $D$-dimensional search space.
A variational autoencoder (VAE) is used mainly for the dimensionality reduction \cite{kingma2013_VAE}.
Although LSBO mitigates the curse of dimensionality, its search performance depends on whether the latent space appropriately reflects the properties of the search space \cite{Tripp2020_W-LBO,Gomez-Bombarelli2018_LSBO,grosnit2021_T-LBO}.
If the limited amount of data available for constructing the latent space is insufficient for VAE training, the BO performance is reduced.
Second, if a long time is required for each biological experiment, the number of iterations needed for optimization cannot be performed within a practical timeframe.
For example, a single testicular tissue culture experiment takes up to 5 weeks \cite{Kamoshita2025_sperm}.
BO often requires dozens of experiments to obtain sufficient improvement; in this case, 50 experiments would take as long as 250 weeks.
In short-term experiments (minutes to days), the total time is often acceptable: BO for 30--100 iterations has been performed in experimental systems where each iteration took only 6--20 min \cite{MacLeod2020,munyebvu2025,nambiar2022}.
Together, high dimensionality and long duration of experiments can hinder the convergence of optimization.

One solution is to augment the dataset with pseudo-experimental data.
Yin et al. \cite{yin2024_TSBO} introduced teacher--student-based semi-supervised learning into LSBO and proposed Teacher--Student Bayesian Optimization (TSBO), in which a teacher model assigns objective values as pseudo-labels to explainable variables sampled in the latent space, and a student model is trained with this information.
By correcting the teacher model predictions through feedback from the student model, TSBO improves the quality of pseudo-experimental data and performs LSBO while updating the surrogate model using a combined dataset of experimental and pseudo-experimental data.
The effectiveness of TSBO depends on how well the latent representation preserves the structure of the search space because TSBO assumes the LSBO framework.
In particular, as mentioned above, insufficient VAE training may degrade BO search performance.
This limitation motivates the development of a method that generates pseudo-experimental data in the search space rather than in the latent space.
Qian et al. \cite{qian2021_Bopp} have proposed a BO method (BOPP) that generates pseudo-experimental data around acquired experimental data and updates the surrogate model using a combined dataset of experimental data and pseudo-experimental data.
Assuming that nearby explainable variables have close objective values, BOPP uses the objective values of experimental data to assign those to the generated nearby pseudo-experimental data.
Because BOPP generates pseudo-experimental data only around experimental data, its effectiveness is limited in regions where the objective function changes drastically.
Thus, pseudo-experimental data--based methods still need substantial improvement for application to small experimental datasets.

In this study, to improve BO performance with small datasets, we propose PolyBO, a BO method that generates pseudo-experimental data from an adaptively updated versatile parametric model retrained whenever new experimental data are acquired.
PolyBO generates pseudo-experimental data in the original search space without relying on latent representations, whose training can become unstable if data are limited.
PolyBO assigns objective values to explainable variables sampled from the search space using a polynomial regression model, and it updates the surrogate model with a combined dataset of experimental and pseudo-experimental data.
PolyBO design was based on the hypothesis that, in small-dataset settings, polynomial regression shortens optimization time, quantified here as the number of optimization iterations.
Polynomial regression is more flexible than a linear model but has lower capacity than deep learning models such as the multilayer perceptron used as the teacher model in TSBO; however, its capacity is sufficient to regress a broad range of functions.
To evaluate whether the PolyBO can reduce the optimization time, we used it to optimize 24 synthetic benchmark functions from Black-Box Optimization Benchmarking (BBOB) \cite{hansen2009} and compared the time required for PolyBO and vanilla BO to reach the same performance.
In high-dimensional setting ($D=20$), PolyBO demonstrated its effectiveness in a high-dimensional setting by achieving a median 42\% reduction in optimization time compared with vanilla BO.
To examine whether PolyBO can be applied to a real-world optimization problem, we used it to optimize composition of a material.
PolyBO demonstrated its effectiveness in this setting by reaching the performance level achieved by conventional methods in a median 96\% shorter optimization time.

\clearpage

\section*{Materials and Methods}

\subsection*{Problem setting}
We consider an optimization problem in which experimental conditions $\mathbf{x}\in \mathcal{X}$ are varied within the experimentally allowable search space $\mathcal{X}\subset \mathbb{R}^{D}$ to search for the parameter $\mathbf{x}^*$ that maximizes the experimental result, represented as the objective function $f:\mathcal{X}\to\mathbb{R}$.

\begin{equation}
    \mathbf{x}^* = \underset{\mathbf{x}} {\operatorname{argmax}} \, f(\mathbf{x})
\end{equation}

\subsection*{PolyBO algorithm}
PolyBO aims to improve the BO efficiency when each experiment requires a long time; pseudo-experimental data generated from polynomial regression accelerate convergence.

\begin{figure}[h]
    \centering
    \includegraphics[width=17cm]{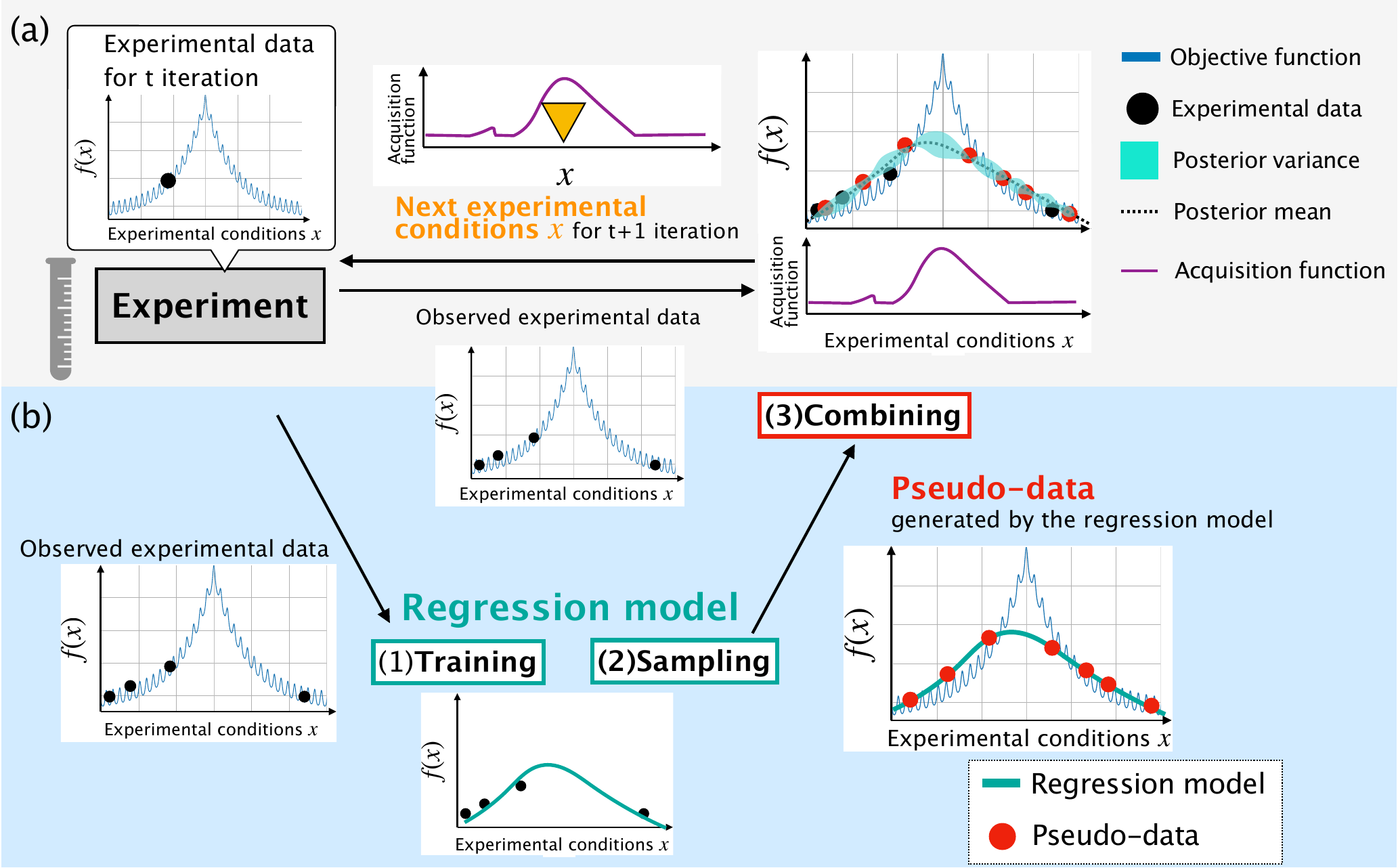}
    \caption{Overview of this study}
    \label{fig:overview}
    \raggedright{
        Schematic illustration of vanilla BO and PolyBO.
        (a) Vanilla BO trains a Gaussian process regression model using only the experimental data up to iteration $t$ and calculates the acquisition function.
        Optimization is then performed by proposing one condition that maximizes the acquisition function, conducting an experiment under that condition, and obtaining the experimental data at iteration $t+1$.
        (b) PolyBO combines vanilla BO with generation of regression-based pseudo-experimental data.
        The regression model is trained using the experimental data up to iteration $t$, and pseudo-experimental data are generated using the trained model.
        The pseudo-experimental data are pairs of explainable variables randomly sampled from the search space, and objective values are calculated by substituting the explainable variables into the regression model.
        The experimental data up to iteration $t$ and the pseudo-experimental data are treated equally and combined into a dataset to train a Gaussian process regression model and calculate the acquisition function.
        Optimization is then performed as above.
    }
\end{figure}

In comparison with vanilla BO (Fig.~\ref{fig:overview}a), three steps are added in PolyBO (Fig.~\ref{fig:overview}b).
In vanilla BO, a Gaussian process regression model is used as a surrogate model and is trained using experimental data.
Using the trained model, an acquisition function is calculated to select the next experimental condition that maximizes the acquisition function, and the next experiment is conducted under the selected condition.
The acquisition function proposes the next experimental conditions by balancing exploration of unobserved regions and exploitation of previously observed objective values.
Expected Improvement (EI) (Eq.~\ref{eq:EI}) \cite{Mockus1975_EI} and GP-UCB (Eq.~\ref{eq:UCB}) \cite{srinivas2010_UCB} have been proposed as representative acquisition functions.
EI calculates the expected improvement over the known best value $f(\mathbf{x}^+)$, whereas GP-UCB calculates an upper confidence bound from the posterior mean $\mu(\mathbf{x})$ and standard deviation $\sigma(\mathbf{x})$ of the trained Gaussian process regression.

\begin{equation}
    \label{eq:EI}
    \alpha_{\text{EI}}(\mathbf{x})=
    \begin{cases}
        (\mu(\mathbf{x})-f(\mathbf{x}^+))\Phi(Z)+\sigma(\mathbf{x})\phi(Z) & \sigma(\mathbf{x})>0 \\
        0                                                                  & \sigma(\mathbf{x})=0
    \end{cases},
    \quad
    Z=\dfrac{\mu(\mathbf{x})-f(\mathbf{x}^+)}{\sigma(\mathbf{x})}
\end{equation}

\begin{equation}
    \label{eq:UCB}
    \alpha_{\text{GP-UCB}}(\mathbf{x}) = \mu (\mathbf{x}) + \sqrt{\beta} \sigma ({\mathbf x})
\end{equation}
Here, $\alpha$ denotes the acquisition function; $\phi$ and $\Phi$ are the probability density function and cumulative distribution function, respectively, of the standard normal distribution; $Z$ is the standardized variable, and $\beta$ is a hyperparameter that controls the exploration--exploitation balance.

In PolyBO, a polynomial regression model is trained using experimental data,
pseudo-experimental data are generated by sampling from the trained polynomial regression model, and the Gaussian process regression model is trained using the combined experimental and pseudo-experimental data,
which are treated equally in the same format as pairs of explainable variables and objective values.
The acquisition function is then calculated as in vanilla BO, and the next experiment is conducted under the condition that maximizes it.

The pseudocode of the PolyBO algorithm is shown in Algorithm \ref{algo:pseudo_experiment_bayesian_optimization}.
At iteration $t=0$, an initial experimental dataset of size $k$, $ \mathcal{D}_{0:0} = \{(\mathbf{x}^{(t=0)}_i, y^{(t=0)}_i)\}_{i=1}^k $, is prepared for training the Gaussian process regression model.
Here, $k$ is the initial dataset size, and the index $i=1,\dots,k$ enumerates the samples in the initial dataset.
The experimental data consist of explainable variables $\mathbf{x}^{(0)}_i$ randomly sampled from the search space and the corresponding objective values $y^{(0)}_i$.
The set consisting of the initial data and the experimental data obtained up to iteration $t-1$ is denoted as $ \mathcal{D}_{0:t-1} = \mathcal{D}_{0:0} \cup \{(\mathbf{x}^{(\tau)}, y^{(\tau)})\}_{\tau=1}^{t-1} $.
Here, $\tau=1,\dots,t-1$ indexes the experimental data sequentially acquired after the initial dataset.
When $t=1$, $\{(\mathbf{x}^{(\tau)}, y^{(\tau)})\}_{\tau=1}^{t-1}$ is the empty set, and therefore $ \mathcal{D}_{0:t-1} = \mathcal{D}_{0:0}$.

The $p^{\mathrm{th}}$-degree polynomial regression model $f_p(\mathbf{x})$ is trained using the experimental data $\mathcal{D}_{0:t-1}$.
Polynomial regression expands the explainable variables $\mathbf{x}$ into a new feature vector $\phi(\mathbf{x})$ and then applies linear regression;
$\phi(\mathbf{x})$ includes not only power terms up to degree $p$ for each explainable variable but also interaction terms,
expressed as products of different explainable variables.
Let $\mathbf{\Phi}$ be the design matrix constructed by stacking $\phi(\mathbf{x})^\top$ for each training sample and $\mathbf{y} = [y_1^{(0)}, \dots, y_k^{(0)}, y^{(1)}, \dots, y^{(t-1)}]^\top$ be the corresponding objective values; the model weight parameter $\mathbf{w}$ is then analytically obtained by solving the normal equation (Eq.~\ref{eq:normal_eq}).
\begin{equation}
    \label{eq:normal_eq}
    \mathbf{w} = (\mathbf{\Phi}^\top \mathbf{\Phi})^{-1} \mathbf{\Phi}^\top \mathbf{y}
\end{equation}
The number of parameters in a model that includes all power and interaction terms is
\[
    \binom{D+p}{p}=\dfrac{(D+1)(D+2)\cdots(D+p)}{p!}.
\]
Pseudo-experimental data $\mathcal{D}_t' = \{(\mathbf{x'}^{(t)}_j, y'^{(t)}_j)\}_{j=1}^{m'} $ are generated using the trained model $f_p(\mathbf{x})$.
Here, $m'$ is the pseudo-experimental dataset size, and $j$ is the sample index.
The pseudo-experimental data consist of explainable variables $\mathbf{x'}^{(t)}_j$ randomly sampled from the search space and objective values $y'^{(t)}_j = f_p(\mathbf{x'}^{(t)}_j)$ calculated by substituting them into the trained model $f_p$.
The posterior distribution is predicted by Gaussian process regression using $\mathcal{D}_{0:t-1} \cup \mathcal{D}_t'$, which combines the pseudo-experimental data $\mathcal{D}_t'$ and experimental data $\mathcal{D}_{0:t-1}$, and then the acquisition function is calculated.
When using EI as the acquisition function, PolyBO uses pseudo-experimental data only as auxiliary information for constructing the Gaussian process posterior and defines the incumbent value $f(\mathbf{x}^+)$ in Eq.~\ref{eq:EI} solely from the experimental data $\mathcal{D}_{0:t-1}$.
This design prevents pseudo-experimental objective values potentially affected by regression error from directly determining the EI improvement baseline.
A limited-memory Broyden--Fletcher--Goldfarb--Shanno algorithm with bound constraints \cite{byrd1995_BFGS} was used to numerically find the maximum of the acquisition function.
The explainable variable $\mathbf{x}^{(t)}$ that maximizes the acquisition function is obtained, and then the experiment is conducted.
Using the experimental result $y^{(t)}$ as the objective value, the experimental data are updated as $\mathcal{D}_{0:t} = \mathcal{D}_{0:t-1} \cup \{(\mathbf{x}^{(t)}, y^{(t)})\}$, and then the iteration proceeds.
PolyBO optimizes experimental conditions by repeating the above process.

At every iteration, the update mechanism in PolyBO replaces the pseudo-experimental dataset with a newly generated dataset $\mathcal{D}_t'$ of size $m'$ and discards pseudo-experimental data generated in the previous iterations.
This prevents low-quality pseudo-experimental data generated when only a few experimental data are available from degrading convergence performance in the middle and later stages of optimization.

\begin{algorithm}
    \caption[Pseudo-experimental-data-assisted BO]{PolyBO algorithm}
    \label{algo:pseudo_experiment_bayesian_optimization}
    \begin{algorithmic}[1]
        \STATE Set initial dataset $\mathcal{D}_{0:0} = \{(\mathbf{x}^{(0)}_i, y^{(0)}_i)\}_{i=1}^k$, where $\mathbf{x}^{(0)}_i$ are randomly sampled from the search space and $y^{(0)}_i$ are obtained from experiments.
        \FOR {$t = 1, 2, \dots$}
        \STATE Train $p^{\mathrm{th}}$-degree polynomial regression model $f_p(\mathbf{x})$ with dataset $\mathcal{D}_{0:t-1}$.
        \STATE Sample $m'$ explainable variables $\{\mathbf{x'}^{(t)}_j\}_{j=1}^{m'}$ randomly from the search space.
        \FOR {$j = 1, 2, \dots, m'$}
        \STATE Compute $y'^{(t)}_j = f_p(\mathbf{x'}^{(t)}_j)$.
        \ENDFOR
        \STATE Set $\mathcal{D}_t' = \{(\mathbf{x'}^{(t)}_j, y'^{(t)}_j)\}_{j=1}^{m'}$.
        \STATE Update the Gaussian process regression model with $\mathcal{D}_{0:t-1} \cup \mathcal{D}_t'$.
        \STATE Compute the acquisition function $\alpha(\mathbf{x})$ using the updated Gaussian process regression model.
        \STATE Select the next point $\mathbf{x}^{(t)} = \operatorname{argmax}_{\mathbf{x}} \alpha(\mathbf{x})$.
        \STATE Conduct an experiment at $\mathbf{x}^{(t)}$ to obtain $y^{(t)}$.
        \STATE Update the dataset $\mathcal{D}_{0:t} = \mathcal{D}_{0:t-1} \cup \{(\mathbf{x}^{(t)}, y^{(t)})\}$.
        \ENDFOR
    \end{algorithmic}
\end{algorithm}

\subsection*{Numerical experiments using synthetic benchmark functions}
To evaluate the performance of PolyBO, we optimized 24 BBOB functions \cite{hansen2009}.
These functions are continuous, with diverse landscapes defined on $[-5, 5]^D$ for arbitrary dimensionality $D$, and they can be regarded as broadly simulating biological phenomena with high-dimensional and complex responses.
These functions were implemented using COmparing Continuous Optimizers (COCO) \cite{hansen2021};
we selected $D=2$ and $5$ as low-dimensional settings and $D=10$ and $20$ as high-dimensional settings.
For each function, we searched for optimal conditions using different initial conditions generated from random numbers based on the Philox algorithm \cite{salmon2011} with 30 random seeds and evaluated the optimization process.
Because PolyBO aims to improve optimization efficiency when the amount of data is small, the initial dataset size was set to two.

Algorithms were compared using simple regret (Eq.~\ref{eq:simple_regret}), defined as the difference between the instance-specific reference optimal value $f(\mathbf{x}^{*})$, predefined by the benchmark, and the maximum value found after $n$ searches.
\begin{equation}
    \text{simple regret} = f(\mathbf{x}^{*}) - \underset{i=1, \;\cdots, \;n}{\text{max}}f(\mathbf{x}^{i})
    \label{eq:simple_regret}
\end{equation}
A smaller simple regret indicates a more optimal value.

EI (Eq.~\ref{eq:EI}) was used as the acquisition function.
All methods were implemented on the basis of BoTorch \cite{botorch}, and the source code and data are available at \url{https://github.com/funalab/PolyBO}.

\subsection*{Performance comparison with vanilla BO}
We compared PolyBO with vanilla BO, random sampling, and BOPP \cite{qian2021_Bopp}, as well as with TSBO \cite{yin2024_TSBO}, the latter only in the highest-dimensional setting ($D=20$).
The maximum number of all searches from the initial conditions was set to 100.

We used the evaluation metric $I_{50}$ \cite{taguchi2026},
which is the number of iterations required for the compared method to reach the simple regret reached by vanilla BO after 50 optimization iterations.
$I_{50}<50$ indicates a shorter optimization time than that of vanilla BO, and small $I_{50}$ indicates high optimization performance.

\subsection*{Analysis of the optimization process}
To compare the optimization process between vanilla BO and PolyBO, the sampled points at each iteration were visualized in a reduced-dimensional space.
As dimensionality-reduction algorithms, we used t-distributed Stochastic Neighbor Embedding (t-SNE) \cite{maaten2008_tsne}, which emphasizes the preservation of the local neighborhood structure, and Uniform Manifold Approximation and Projection (UMAP) \cite{mcinnes2018_umap}, which tends to preserve global structures such as inter-cluster distances more accurately than t-SNE.
To sufficiently cover the entire high-dimensional search space, 40,000 explainable variables were obtained by Latin hypercube sampling \cite{Liang2025_lhs}, and dimensionality was reduced on these variables together with all sampled points from vanilla BO and PolyBO.

We also quantified optimization behavior to support the dimensionality reduction--based assessment.
As evaluation metrics, we used Observation Entropy (OE) (Eq.~\ref{eq:OE}) \cite{papenmeier2025_OE} and Proposed-point Entropy (PE) (Eq.~\ref{eq:PE}), which we devised, and compared them between vanilla BO and PolyBO.

\begin{equation}
    \label{eq:OE}
    \mathrm{OE}_{t}
    =
    \frac{D}{t}\sum_{i=1}^{t}\log \epsilon_i^{k}
    + \psi(t)
    - \psi(1)
    + \log V_{D}
\end{equation}

\begin{equation}
    \label{eq:PE}
    \mathrm{PE}_{t}
    =
    -\int p(y\mid \mathbf{x}^{(t)}, \mathcal{D}_{0:t-1} \cup \mathcal{D}_{t}')
    \log p(y\mid \mathbf{x}^{(t)}, \mathcal{D}_{0:t-1} \cup \mathcal{D}_{t}')\,dy
    =
    \frac{1}{2}\log\left(2\pi e\,\sigma_t^2(\mathbf{x}^{(t)})\right).
\end{equation}
Here, $\epsilon_i^{k}$ is the distance from point $\mathbf{x}^{(i)}$ to its $k^{\mathrm{th}}$ nearest neighbor, $V_{D}=\pi^{D/2}/\Gamma(1+D/2)$ is the volume of a $D$-dimensional unit ball, $\Gamma$ is the gamma function, and $\psi$ is the digamma function.
Following Papenmeier et al. \cite{papenmeier2025_OE}, we set $k=\log(t)$.

OE quantifies the density of explainable variables among the sampled points obtained up to iteration $t$, whereas PE quantifies the predictive variance at the explainable variable proposed at iteration $t$.
In BO, it is important to consider both explainable variables and objective values.
OE focuses on explainable variables and PE on objective values, i.e. these metrics quantify complementary aspects of uncertainty.
For both metrics, higher values indicate a more exploration-oriented method.

\subsection*{Robustness analysis of hyperparameters and acquisition functions}
The PolyBO-specific hyperparameters are the pseudo-experimental dataset of size $m'$ and the maximum polynomial degree $p$.
We assessed how changes in the hyperparameters affect PolyBO performance (sensitivity analysis).
At $D=20$, one hyperparameter was fixed, while the other was varied.
We measured $I_{50}$ at $p = 4$ and $m'$ varied over $\{1,\; 5,\; 10,\; 15,\; 20,\; 25,\; 40,\; 80,\; 100,\; 200\}$, and at $m' = 10$ and $p$ varied over $\{1,\; 2,\; 3,\; 4,\; 5,\; 6,\; 7,\; 8\}$.

To examine whether the optimization performance of PolyBO depends on the EI acquisition function (Eq.~\ref{eq:EI}), we compared the optimization process with that obtained using GP-UCB (Eq.~\ref{eq:UCB}).
The hyperparameter $\beta$ affects the optimization strategy and was fixed at $\beta=1$ as in \cite{fromer2025_beta1,zhao2024_beta1}; it was also used by Papenmeier et al. \cite{papenmeier2025_OE} because $\beta$ has no marked bias toward exploratory or exploitative behavior.

\subsection*{Ablation study of the update mechanism}
The update mechanism of PolyBO (i) discards past pseudo-experimental data and (ii) maintains a constant pseudo-experimental dataset size.
To clarify the effect of the former on optimization performance, we evaluated how a mechanism that does not discard pseudo-experimental data and uses all such previously generated data (the without-reset mechanism) affects optimization performance.
To clarify the effect of the latter, we evaluated the $m'$-scaling mechanism, which does not retain pseudo-experimental data from previous iterations but instead generates a fresh pseudo-experimental dataset of size $t \times m'$ at iteration $t$.

\subsection*{Application to a real-world material composition optimization problem}
Simulated material composition results generated by a neural-network predictor for high-entropy alloys (alloys formed by mixing five or more elements in similar proportions; HEAs) \cite{xian2025_RL,wang2023_HEA_NN} were used as a proxy for experimental results.
This simulation model maps composition parameters of up to 10 elements to key mechanical properties: yield strength ($\sigma_Y$), ultimate tensile strength ($\sigma_U$), and elongation ($\varepsilon$).
Although HEAs are not the biomaterials that are the main target of this study, we selected them as a representative example of an expensive-to-evaluate optimization problem because, like biomaterials, their synthesis requires a long time.

We optimized composition parameters for 10 elements.
For evaluation, we used the HEA figure of merit ($\rm{FOM_{HEA}}$) (Eq.~\ref{eq:FOM_HEA}).
$\rm{FOM_{HEA}}$ represents the inherent strength--ductility trade-off as a weighted combination of the three normalized properties.
As in Xian et al. \cite{xian2025_RL}, we evaluated $\rm{FOM_{HEA}}$ against Experimental Iteration, which is the number of actual experiments performed.
\begin{equation}
    \label{eq:FOM_HEA}
    {{\rm{FOM}}}_{{\rm{HEA}}}=\frac{1}{3}\left(\frac{{\sigma }_{Y}}{{{\sigma }_{Y}}_{N}}+\frac{{\sigma }_{U}}{{{\sigma }_{U}}_{N}}+\frac{\varepsilon }{{\varepsilon }_{N}}\right)
\end{equation}
The subscript $N$ indicates the normalized value of the corresponding property.

We compared PolyBO with BO-EI and the reinforcement-learning method RL-DQN \cite{xian2025_RL}.
A search for optimal conditions using different initial conditions was based on 30 random seeds (the HEA NN predictor for PolyBO and BO-EI) or 24 random seeds (RL-DQN).
We used the same 20 initial conditions for PolyBO as for BO-EI.
The upper limit of the number of searches was set to 1500, including the initial data.
EI (Eq.~\ref{eq:EI}) was used as the acquisition function.

Because this was a discrete and constrained optimization problem, we extended PolyBO to handle this search space (Supplementary Algorithm \ref{algo:pseudo_experiment_discrete_bo}).
Following BO-EI, we developed an algorithm in which the atomic fraction $x_i$ of each element is restricted to a discrete grid with step size $\delta$ and simultaneously satisfies the equality constraint $\sum_{i=1}^{d} x_i = 1$ as well as the lower and upper bound constraints $l_i \le x_i \le u_i$ for each element.
We generated pseudo-experimental data under the constraint that the sampled explainable variables were drawn uniformly from the discrete grid $\mathcal{G}_\delta$, which satisfied the above constraints.
This sampling scheme generated pseudo-experimental data in the same space as experimental data and was expected to provide a more appropriate training-data distribution for Gaussian process regression than sampling from the corresponding continuous constrained space.

We quantified PolyBO performance using the number of iterations it required to reach the $\rm{FOM_{HEA}}$ attained by RL-DQN after 1500 optimization iterations ($I_{1500}$).

\clearpage

\section*{Results}
\subsection*{Performance of PolyBO on synthetic benchmark functions}
\begin{figure}[h]
    \centering
    \includegraphics[width=17cm]{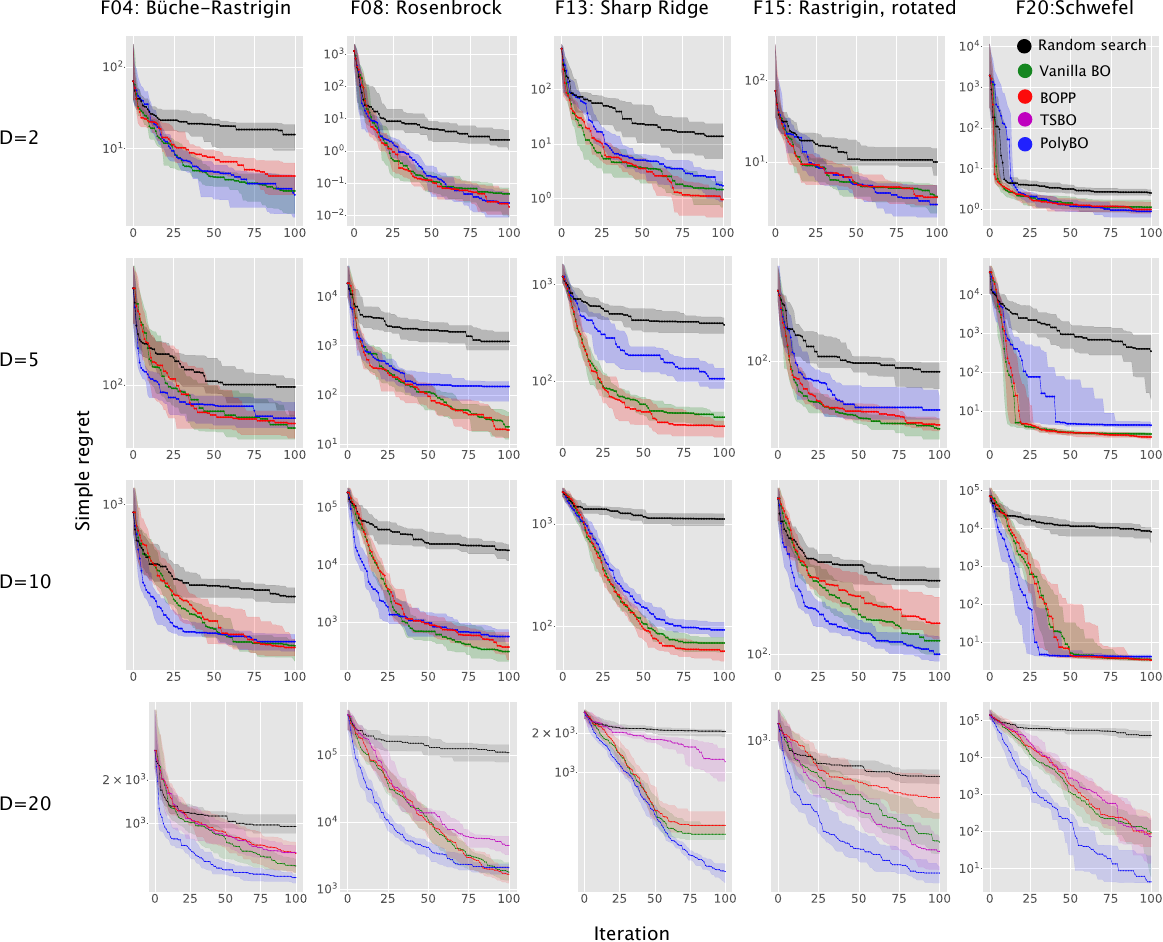}
    \caption{Optimization processes using PolyBO and conventional methods}
    \label{fig:result1}
    \raggedright{
        The median (lines) and interquartile range (shading) of 30 runs of optimization from different initial conditions are shown for each method.
        D, dimensions; function IDs (F) are shown for 5 functions selected from the 24 functions.
        $m'=10$, $p=4$; TSBO was used only at $D=20$.
    }
\end{figure}

To evaluate PolyBO, we compared it with conventional BO, random sampling, and BOPP, which generates pseudo-experimental data around acquired experimental data \cite{qian2021_Bopp}.
At $D=20$, we also compared PolyBO with TSBO, which incorporates semi-supervised learning \cite{yin2024_TSBO}.
Across the 24 benchmark functions, PolyBO tended to achieve lower simple regret earlier than vanilla BO at $D=10$ and $D=20$, but not at $D=2$ or $D=5$ (Fig.~\ref{fig:result1}, Supplementary Figs.~\ref{fig:supp_benchmark_d2}--\ref{fig:supp_benchmark_d20}).

At $D=20$, PolyBO decreased $I_{50}$ for all benchmark functions except F05, F13, F16, F21, and F23.
This indicated that PolyBO required fewer iterations for optimization (Table~\ref{tb:result_quantitative}).
The median rate of increase was 12\% at $D=2$, whereas the median rates of reduction were 6\% at $D=5$, 31\% at $D=10$, and 42\% at $D=20$.
At $D=5$, the value could not be calculated for any benchmark functions other than F04, F05, F06, and F16 because the reference simple regret was not reached within 100 iterations.
These results indicated that the optimization performance of PolyBO was low in low-dimensional settings, particularly at $D=5$, but was high in high-dimensional settings.

\subsection*{Analysis of the optimization process}
\begin{figure}[h]
    \centering
    \includegraphics[width=16cm]{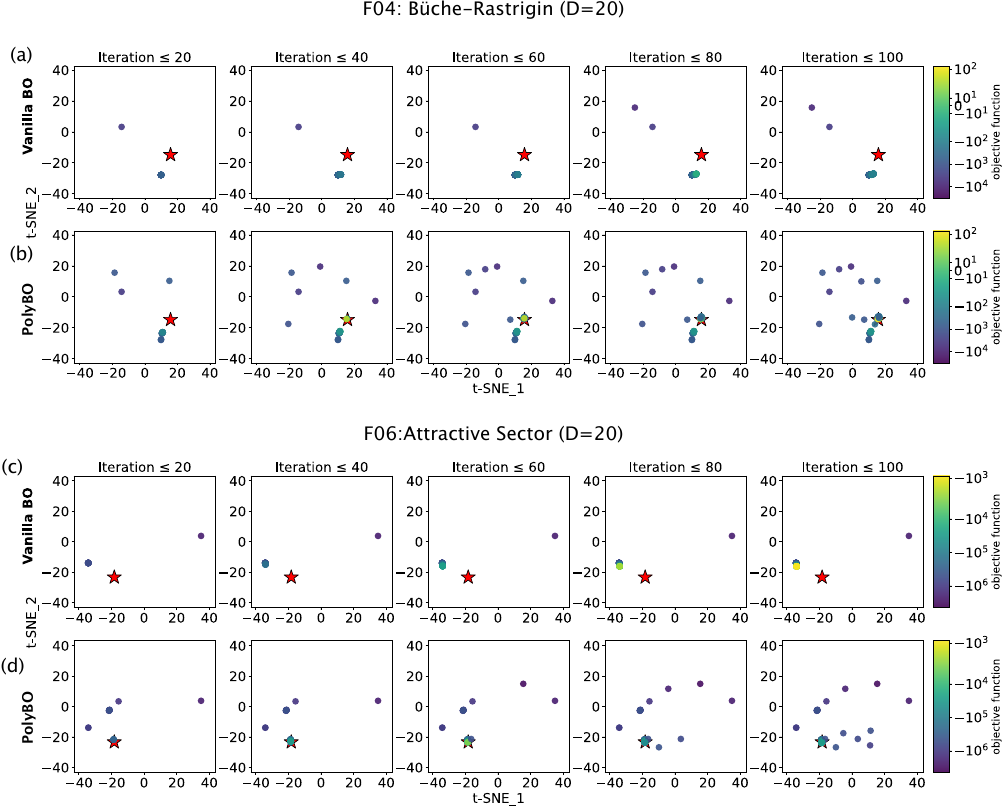}
    \caption{t-SNE visualization of optimization processes for F04 and F06 at $D=20$}
    \label{fig:tsne}
    \raggedright{
        Circles indicate all sampled points used in the optimization, including the initial experimental points and the candidate points $\mathbf{x}^{(t)}$ selected by each method at subsequent iterations.
        Color indicates the value of the objective function.
        A symlog scale was used for the color bar, and the threshold for the linear region was set to $10^{-3}$ of the maximum absolute plotted value of the objective function.
        Red stars indicate the locations of the global optima.
        All results are shown for $m'=10$ and $p=4$.
    }
\end{figure}

We visualized optimization processes for different functions at various dimensionalities using scatter plots.
As representative examples, the optimization processes for F04 and F06 ($D=20$) are shown in Figure~\ref{fig:tsne} (t-SNE) and Supplementary Figures~\ref{fig:supp_umap_d20} (UMAP), \ref{fig:supp_tsne_d2} ($D=2$), and \ref{fig:supp_tsne_d5} ($D=5$).
The entire search spaces for F04 and F06 are shown in Supplementary Fig.~\ref{fig:supp_tsne_landscape}.
At $D=20$, the tendency of PolyBO to explore a broader region of the search space than vanilla BO (Fig.~\ref{fig:tsne} and Supplementary Fig.~\ref{fig:supp_umap_d20}) indicated that PolyBO is an exploration-oriented method.
The t-SNE and UMAP results showed little difference in the global trends, supporting this interpretation.
At $D=2$ and $D=5$, little difference was found in the extent of the search space covered by PolyBO and vanilla BO (Supplementary Figs.~\ref{fig:supp_tsne_d2} and \ref{fig:supp_tsne_d5}).
At $D=20$, PolyBO found promising regions near the global optimum between iterations 20 and 40 for F04 and between iterations 0 and 20 for F06.

\begin{figure}[h]
    \centering
    \includegraphics[width=16cm]{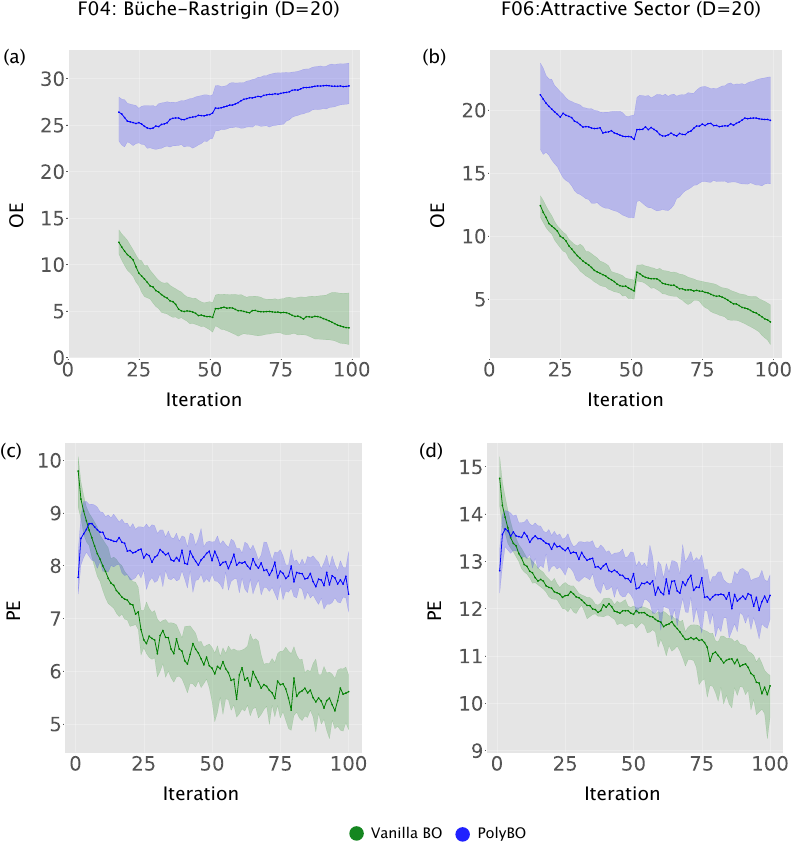}
    \caption{Quantification of optimization behavior for F04 and F06 at $D=20$ using OE and PE}
    \label{fig:entropy20}
    \raggedright{
        The median (lines) and interquartile range (shading) of 30 runs of optimization from different initial conditions are shown for each method.
        $m'=10$, $p=4$.
    }
\end{figure}

We quantified optimization behavior using OE for uncertainty in the input space and PE for predictive uncertainty in objective values (Fig.~\ref{fig:entropy20}, Supplementary Figs.~\ref{fig:supp_entropy_d2} and \ref{fig:supp_entropy_d5}).
At $D=2$, $5$, and $20$, both metrics showed that PolyBO was more exploration-oriented than vanilla BO.
These quantitative results supported the qualitative results from the search-space visualization for high-dimensional settings but not for low-dimensional settings.

\subsection*{Robustness analysis of hyperparameters and acquisition functions}
\begin{figure}[h]
    \centering
    \includegraphics[width=12cm]{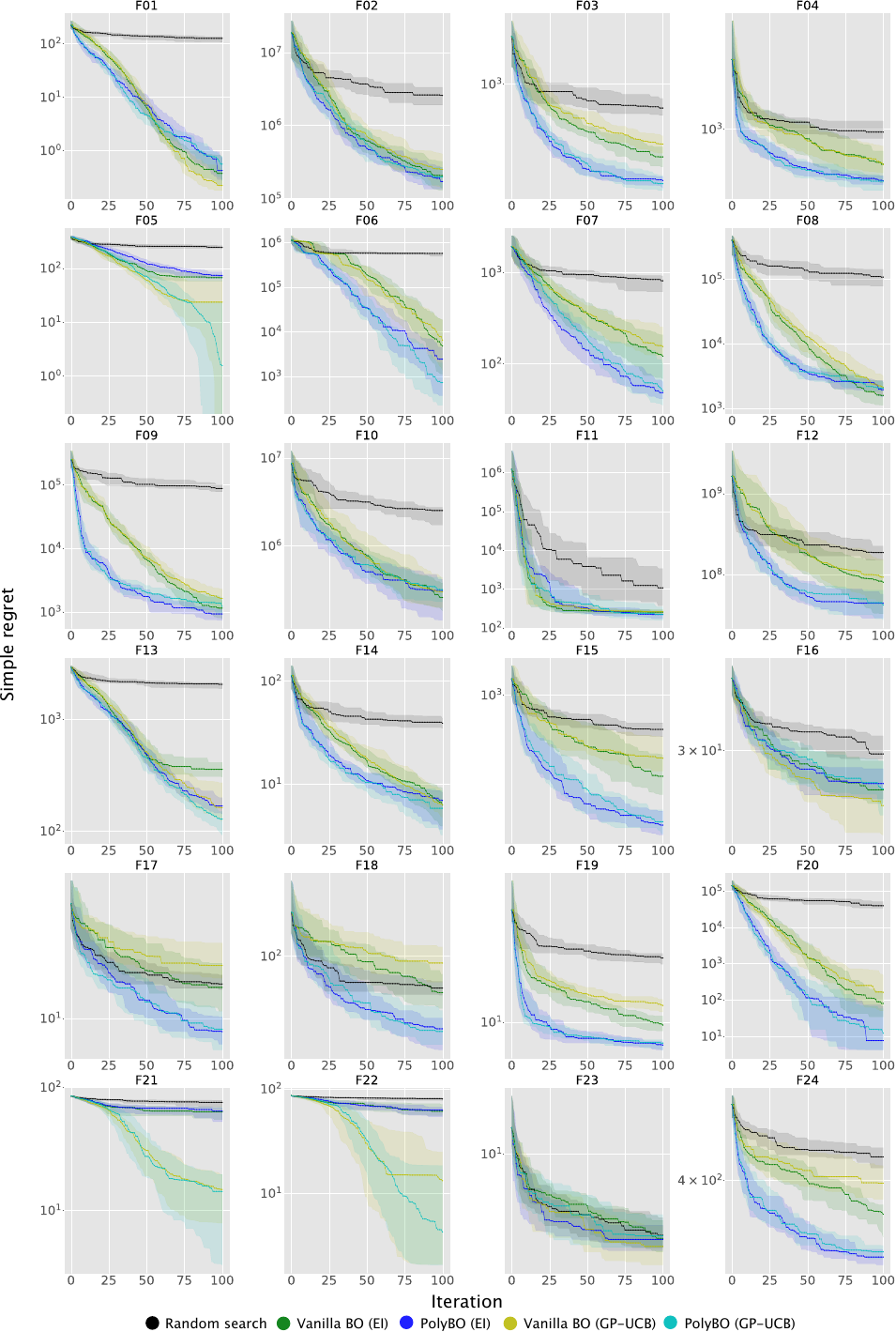}
    \caption{Sensitivity analysis of the acquisition function in PolyBO}
    \label{fig:UCB}
    \raggedright{
        EI or GP-UCB was used as the acquisition function.
        The median (lines) and interquartile range (shading) of 30 runs of optimization from different initial conditions are shown for each method.
        $D=20$, $m'=10$, and $p=4$.
    }
\end{figure}

To evaluate the robustness of the PolyBO-specific hyperparameters $m'$ and $p$, we assessed the effects of their variation on PolyBO performance through sensitivity analysis (Tables~\ref{tb:sensitivity_mprime} and \ref{tb:sensitivity_p}).
To examine whether the optimization performance of PolyBO depends on the EI acquisition function, we compared it with that obtained using GP-UCB (Fig.~\ref{fig:UCB}).

At a maximum of 100 iterations, PolyBO had better optimization performance than vanilla BO over the tested range of pseudo-experimental dataset sizes up to 200 (Table~\ref{tb:sensitivity_mprime}).
This suggests that PolyBO is unlikely to underperform vanilla BO even when the pseudo-experimental dataset size is increased.
Optimization was maximized at $m'=5\text{--}25$.
This suggests a nonlinear relationship between optimization performance and $m'$.

At $p\neq1$, PolyBO had better optimization performance than vanilla BO (Table~\ref{tb:sensitivity_p}).
The median optimization performance, ranked from best to worst, was achieved at $p=4,\;8,\;6,\;7,\;2,\;5,\;3,\;1$.
The suggestion was that even $p$ values $(2,\;4,\;6,\;8)$ tended to result in higher optimization performance than the odd values $(1,\;3,\;5,\;7)$.

For benchmark functions other than F05, F21, and F22, vanilla BO (EI) and vanilla BO (GP-UCB) showed similar trends, as did PolyBO (EI) and PolyBO (GP-UCB) (Fig.~\ref{fig:UCB}).
For F05, F21, and F22, vanilla BO (GP-UCB) outperformed vanilla BO (EI), and PolyBO (GP-UCB) outperformed PolyBO (EI); PolyBO (GP-UCB) optimization performance was comparable to or higher than that of vanilla BO (GP-UCB).
These results indicate that the performance improvement of PolyBO depends little on differences in acquisition functions.

\subsection*{Ablation study of the update mechanism}
We analyzed the two components of the update mechanism, namely discarding pseudo-experimental data and maintaining constant $m'$ (Fig.~\ref{fig:forget}).
Optimization performance was higher in the model with the update mechanism than in those with the without-reset mechanism or the $m'$-scaling mechanism, which does not maintain constant $m'$; the difference was greater in the former than in the latter case.
These results show that both discarding pseudo-experimental data and maintaining constant $m'$ are important and contribute to performance improvement, especially through the update mechanism that discards pseudo-experimental data generated in previous iterations and updates the pseudo-experimental dataset at each iteration.

\subsection*{Application to a real-world material composition optimization problem}
\begin{figure}[h]
    \centering
    \includegraphics[width=16cm]{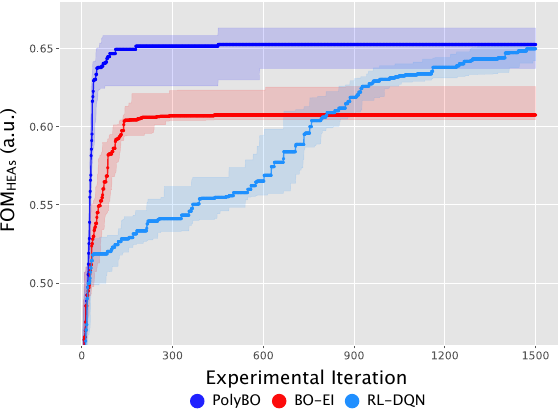}
    \caption{Real-world material composition optimization with PolyBO and two conventional methods}
    \label{fig:FOM_HEA}
    \raggedright{
        The median (lines) and interquartile range (shading) are shown.
        The results of 30 (PolyBO and BO-EI) or 24 (RL-DQN) independent runs with different initial conditions are shown.
        The initial conditions in PolyBO were the same 20 conditions as used in BO-EI by Xian et al. \cite{xian2025_RL}.
        The results of BO-EI and RL-DQN were obtained by reproducing the published code \cite{xian2025_RL}.
        $D=10$, $m'=10$, and $p=4$.
    }
\end{figure}

We optimized material composition and compared the optimization processes (Fig.~\ref{fig:FOM_HEA}).
The optimization time was shorter for PolyBO than for the conventional methods.

The median value of the metric $I_{1500}$ was 59.
This indicated that PolyBO reached the performance level achieved by RL-DQN with a 96\% shorter optimization time.
This result suggests that PolyBO can accelerate optimization when applied not only to synthetic benchmark functions but also to experimentally grounded, real-world search spaces.

\clearpage

\section*{Discussion}
We developed PolyBO, an optimization method that uses regression-based pseudo-experimental data to reduce the optimization time.
In PolyBO, BO is conducted with augmented data from an adaptively updated versatile parametric model, and these pseudo-experimental data accelerate convergence even when each experiment requires a long time (Fig.~\ref{fig:overview}).
In high-dimensional search spaces, PolyBO found more optimal parameters in fewer iterations than did conventional methods (Figs.~\ref{fig:result1} and \ref{fig:FOM_HEA}) and reached comparable conditions in fewer iterations (Table~\ref{tb:result_quantitative}).
These results demonstrate that PolyBO improves optimization performance in high-dimensional settings.
Although this study did not directly evaluate biomaterials, PolyBO may also be applicable to time-consuming biological experiments.
By reducing the optimization time, PolyBO can enable searches for experimental conditions that were previously difficult to conduct in time-consuming experimental systems, such as testicular tissue culture, where a single experiment takes up to 5 weeks \cite{Kamoshita2025_sperm}.

Visualization of the optimization process showed that PolyBO found promising regions near the global optimum by emphasizing exploration in the early phase (Fig.~\ref{fig:tsne}).
The high performance of PolyBO on the F04 and F06 functions (Table~\ref{tb:result_quantitative}, Supplementary Fig.~\ref{fig:supp_benchmark_d20}) can be attributed to its exploration-oriented behavior that allowed it to find promising regions earlier than did vanilla BO.
Thus, in high-dimensional settings with broad search spaces, the stronger exploratory behavior of PolyBO may have improved optimization performance by enabling earlier discovery of promising regions.

In low-dimensional settings, the discrepancy between the qualitative results from scatter-plot visualization of the optimization process (Fig.~\ref{fig:tsne}) and the quantitative results from OE and PE (Fig.~\ref{fig:entropy20}) arises from the different aspects of optimization behavior:
the scatter-plots show the search range of sampled points, whereas OE evaluates density from inter-point distances, and PE evaluates the predictive uncertainty of proposed points.
In low-dimensional settings, the search space is relatively small, so vanilla BO can cover a broad search range with fewer sampled points than in high-dimensional settings.
As a result, differences between PolyBO and vanilla BO are not obvious in the scatter-plots.
However, even when the search range of sampled points is comparable between PolyBO and vanilla BO, PolyBO may select sparse regions far from existing sampled points or points with high predictive uncertainty more frequently than vanilla BO, and these differences may have been reflected in OE and PE.

Sensitivity analysis provided guidance for selecting the maximum polynomial degree $p$.
When $p=1$, the convergence performance of PolyBO was low (Table~\ref{tb:sensitivity_p}).
This was likely because the polynomial regression model has only 21 parameters and low model expressiveness when $p=1$, preventing the regression model from capturing complex benchmark-function landscapes.
When $p$ was varied from 1 to 8, the only benchmark function for which $p=1$ showed the highest optimization performance was F05, whose search space is mostly linear (Table~\ref{tb:sensitivity_p}).
A linear regression model ($p=1$) is advantageous when the response of the objective function to the explainable variables is expected to be linear, whereas nonlinear regression models ($p\neq1$) are advantageous for nonlinear responses.

For F08, F09, F10, and F20, optimization performance was equal to or higher at even degrees ($p=2,\;4,\;6,\;8$) than at odd degrees ($p=1,\;3,\;5,\;7$) (Table~\ref{tb:sensitivity_p}).
F08 and F09 have bent valleys, F10 has a curvature structure with strong ill-conditioning, F20 has a complex global structure near the boundary \cite{hansen2009}, and pseudo-experimental data generated by even-degree polynomials may be more advantageous for optimization than asymmetric pseudo-experimental data that can arise from odd-degree polynomials.

\begin{figure}[h]
    \centering
    \includegraphics[width=12cm]{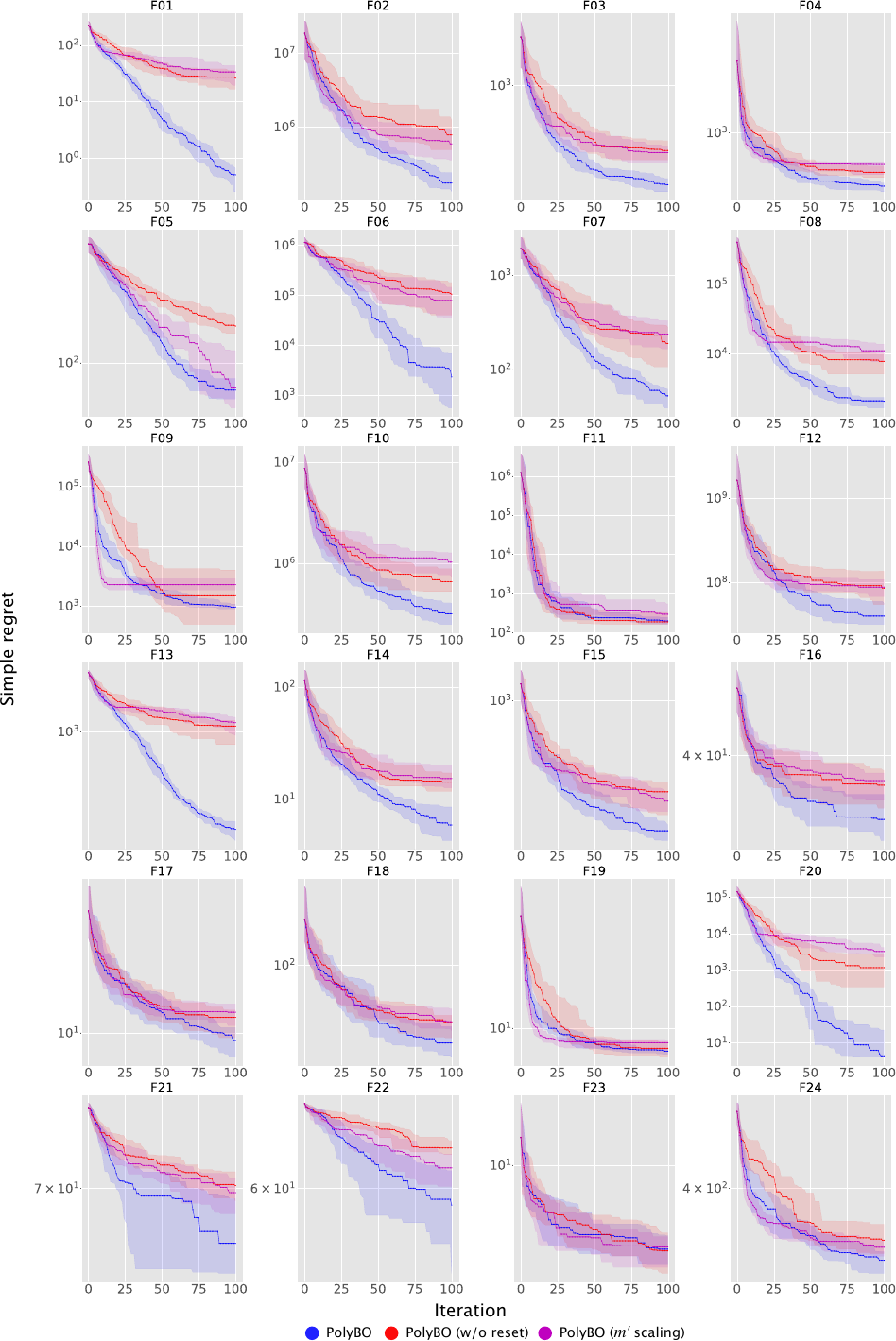}
    \caption{Ablation study of the update mechanism}
    \label{fig:forget}
    \raggedright{
        The update mechanism was ablated by either not discarding previously generated pseudo-experimental data (w/o reset) or increasing the pseudo-experimental dataset size ($m'$-scaling).
        The median (lines) and interquartile range (shading) of 30 runs of optimization from different initial conditions are shown for each PolyBO variant.
        $D=20$, $m'=10$, and $p=4$.
    }
\end{figure}

In the ablation study, optimization performance was lower with the without-reset mechanism and with the $m'$-scaling mechanism than with the update mechanism.
The data in Table~\ref{tb:sensitivity_mprime} supported this result: optimization performance was maximized when the pseudo-experimental dataset size was 5--25; because the $m'$-scaling mechanism increases this size to $t \times m'$ at iteration $t$, it exceeds this favorable range after the early iterations when $m'=10$, explaining the reduced optimization performance shown in Fig.~\ref{fig:forget}.

In low-dimensional settings, particularly at $D=5$, PolyBO optimization performance was low.
The suggestion was that it depends on the balance between the size of the search space and the regression error of the pseudo-experimental data.
At $D=2$, the exploration-enhancing property of PolyBO may not have contributed substantially because the search space was small, and vanilla BO could easily reach promising regions.
In low-dimensional settings, the expectation that the regression error of the polynomial regression model will be relatively small makes it less likely that pseudo-experimental data strongly impede optimization; consequently, PolyBO performance remained comparable to that of vanilla BO or was lower only in limited cases.
At $D=5$, vanilla BO may already have sufficient convergence performance, whereas its broader search space than at $D=2$ makes the influence of pseudo-experimental data non-negligible.

To evaluate the memory complexity of the PolyBO algorithm, we measured peak memory usage required to solve the normal equation (Eq.~\ref{eq:normal_eq}) while increasing the maximum degree $p$ from 1 to 9 at $D=20$, and we observed exponential growth (Supplementary Fig.~\ref{fig:supp_memory}a).
As the maximum degree $p$ increases, the number of parameters increases as $O\binom{D+p}{p}$ (Supplementary Fig.~\ref{fig:supp_memory}b), and the memory usage for solving the normal equation (Eq.~\ref{eq:normal_eq}) in polynomial regression theoretically increases as $O\left(\binom{D+p}{p}^2\right)$.
Although the optimal maximum degree differs among benchmark functions (Table~\ref{tb:sensitivity_p}), increasing the maximum degree $p$ exponentially increases memory complexity, and the optimal degree must therefore be determined while considering hardware specifications.

Our data indicate that an optimal pseudo-experimental dataset size rather than addition of a large amount of pseudo-experimental data (Table~\ref{tb:sensitivity_mprime}) and its sequential updating are important for maximizing optimization performance (Fig.~\ref{fig:forget}).
It is thus important to incorporate the low-fidelity source provided by the polynomial regression model into the surrogate model with an appropriate strength and to update this source as iterations proceed.
This approach is similar to that used in multi-fidelity Bayesian optimization (MFBO) \cite{kandasamy2019_MFBObase,Foumani2023_MFBO2,Do2025_MFBOreview}, which improves the efficiency of optimizing a high-fidelity objective function by using multiple information sources with different evaluation costs and accuracies.
If the pseudo-experimental dataset size is too small, enhancement of exploration is insufficient; if it is too large, there is a risk that the surrogate model becomes biased by over-reliance on pseudo-experimental data.
In future work, it will be important to extend PolyBO to a framework that sequentially controls the influence of pseudo-experimental data according to regression uncertainty and consistency with experimental data, rather than fixing the size of the pseudo-experimental dataset.
Because the fidelity of pseudo-experimental data in PolyBO changes as iterations proceed and many existing MFBO methods assume fixed fidelity for each information source \cite{kandasamy2019_MFBObase,Foumani2023_MFBO2,Do2025_MFBOreview}, extending MFBO to model iteration-dependent fidelity will be an important direction for future work.

\clearpage

\bibliographystyle{unsrt}
\bibliography{references}

\section*{Code availability}
The source code of this study is available from \url{https://github.com/funalab/PolyBO}.

\section*{Acknowledgements}
The research was funded by JST CREST, Japan Grant Number JPMJCR21N1 to A.F.
We are grateful to Asst. Prof. Yuki TSUKADA for helpful discussions.

\section*{Author contributions}
\textbf{H. S. :} Data curation, Formal analysis, Investigation, Methodology, Software, Validation, Visualization, Writing--original draft, Writing--review \& editing.
\textbf{Y. T. :} Formal analysis, Methodology, Software, Visualization.
\textbf{K. M. :} Formal analysis, Methodology, Software.
\textbf{Y. H. :} Formal analysis, Writing--review \& editing.
\textbf{T. M. :} Conceptualization, Formal analysis, Methodology, Supervision, Visualization, Writing--review \& editing.
\textbf{A. F. :} Conceptualization, Formal analysis, Funding acquisition, Methodology, Project administration, Resources, Supervision, Visualization, Writing--review \& editing.

\section*{Competing Interests}
The authors declare no competing interests.

\clearpage

\clearpage

\clearpage
\begin{table}[b]
    \begin{center}
        \caption{The number of iterations required for optimization by PolyBO and vanilla BO}
        \label{tb:result_quantitative}
        \vspace{1em}
        \begin{tabular}{
                ll
                D{(}{\,(}{3}
                D{(}{\,(}{3}
                D{(}{\,(}{3}
                D{(}{\,(}{3}
            }
            \hline
            \multicolumn{1}{l}{ID} &
            \multicolumn{1}{l}{Benchmark function} &
            \multicolumn{1}{c}{$D=2$} &
            \multicolumn{1}{c}{$D=5$} &
            \multicolumn{1}{c}{$D=10$} &
            \multicolumn{1}{c}{$D=20$} \\
            \hline
            F01 & Sphere & \bm{17}(8) & $\textendash$ & $\textendash$ & \bm{49}(12) \\
            F02 & Separable ellipsoidal & $\textendash$ & $\textendash$ & $\textendash$ & \bm{39}(18) \\
            F03 & Rastrigin, original & \bm{48}(58) & $\textendash$ & \bm{27}(22) & \bm{23}(9) \\
            F04 & B{\"u}che--Rastrigin & 65($\textendash$) & 76($\textendash$) & \bm{30}(38) & \bm{19}(19) \\
            F05 & Linear Slope & \bm{8}(7) & \bm{16}(9) & \bm{44}(13) & 74($\textendash$) \\
            F06 & Attractive sector & 87($\textendash$) & \bm{24}(29) & \bm{35}(33) & \bm{28}(16) \\
            F07 & Step ellipsoidal & \bm{32}(52) & $\textendash$ & 71($\textendash$) & \bm{29}(14) \\
            F08 & Rosenbrock, original & 59(18) & $\textendash$ & 73($\textendash$) & \bm{26}(7) \\
            F09 & Rosenbrock, rotated & 57(25) & $\textendash$ & \bm{27}(28) & \bm{22}(13) \\
            F10 & Ellipsoidal & 91($\textendash$) & $\textendash$ & 68($\textendash$) & \bm{39}(34) \\
            F11 & Discus & 90($\textendash$) & $\textendash$ & \bm{41}(66) & \bm{45}(58) \\
            F12 & Bent cigar & \bm{46}(34) & $\textendash$ & $\textendash$ & \bm{24}(14) \\
            F13 & Sharp ridge & 67(45) & $\textendash$ & 65($\textendash$) & 50(9) \\
            F14 & Different Powers & 62(37) & $\textendash$ & 72($\textendash$) & \bm{30}(11) \\
            F15 & Rastrigin, rotated & \bm{48}(36) & $\textendash$ & \bm{28}(21) & \bm{13}(13) \\
            F16 & Weierstrass & $\textendash$ & 70($\textendash$) & 64($\textendash$) & 94($\textendash$) \\
            F17 & Schaffer's F7 & 59(48) & $\textendash$ & \bm{32}(20) & \bm{15}(21) \\
            F18 & Schaffer's F7, moderately ill-conditioned & $\textendash$ & $\textendash$ & \bm{26}(21) & \bm{16}(20) \\
            F19 & Composite Griewank--Rosenbrock function F8F2 & \bm{31}(44) & $\textendash$ & \bm{33}(37) & \bm{10}(9) \\
            F20 & Schwefel & \bm{42}(46) & $\textendash$ & \bm{31}(46) & \bm{30}(11) \\
            F21 & Gallagher's Gaussian 101-me peaks & 54(58) & $\textendash$ & \bm{36}($\textendash$) & 74($\textendash$) \\
            F22 & Gallagher's Gaussian 21-hi peaks & 56(43) & $\textendash$ & 52($\textendash$) & \bm{47}(53) \\
            F23 & Katsuura & 58(75) & $\textendash$ & \bm{26}(36) & 63($\textendash$) \\
            F24 & Lunacek bi-Rastrigin & \bm{47}(62) & $\textendash$ & \bm{26}(21) & \bm{12}(16) \\
            \hline\hline
            \multicolumn{2}{c}{Median} &
            \multicolumn{1}{c}{56.0} &
            \multicolumn{1}{c}{47.0} &
            \multicolumn{1}{c}{34.5} &
            \multicolumn{1}{c}{29.0} \\
            \hline \hline
        \end{tabular}
    \end{center}
    \vspace{1em}
    \raggedright{
        To quantify the reduction in optimization time, we calculated the number of iterations required for PolyBO to reach the simple regret reached by vanilla BO at 50 iterations.
        Each value is the median number of iterations at which PolyBO reached the reference simple regret, and the values in parentheses indicate the interquartile range.
        Values $<50$ are in bold, and values $\geq50$ are in plain font.
        The maximum number of iterations was set to 100; \textendash{} indicates that the value could not be calculated because the reference simple regret was not reached within 100 iterations.
        $m'=10$, $p=4$.
    }
\end{table}

\clearpage

\clearpage

\clearpage
\begin{table}[b]
    \begin{center}
        \footnotesize
        \caption{Sensitivity analysis of the pseudo-experimental dataset size $m'$ in PolyBO}
        \label{tb:sensitivity_mprime}
        \vspace{1em}
        \begin{tabular}{l *{10}{D{(}{\,(}{3}}}
            \hline
            \multicolumn{1}{l}{ID} &
            \multicolumn{1}{c}{$m'=1$} &
            \multicolumn{1}{c}{$m'=5$} &
            \multicolumn{1}{c}{$m'=10$} &
            \multicolumn{1}{c}{$m'=15$} &
            \multicolumn{1}{c}{$m'=20$} &
            \multicolumn{1}{c}{$m'=25$} &
            \multicolumn{1}{c}{$m'=40$} &
            \multicolumn{1}{c}{$m'=80$} &
            \multicolumn{1}{c}{$m'=100$} &
            \multicolumn{1}{c}{$m'=200$} \\
            \hline
            F01 & 53(8) & 52(14) & \bm{49}(12) & 56(10) & 50(10) & 59(22) & 57(25) & 97($\textendash$) & $\textendash$ & $\textendash$ \\
            F02 & \bm{43}(20) & \bm{46}(17) & \bm{39}(18) & \bm{41}(24) & \bm{42}(24) & \bm{42}(26) & \bm{42}(25) & 55(40) & 58(46) & 60(51) \\
            F03 & \bm{31}(17) & \bm{29}(17) & \bm{23}(9) & \bm{25}(16) & \bm{23}(7) & \bm{25}(8) & \bm{29}(11) & \bm{28}(18) & \bm{34}(27) & \bm{42}(43) \\
            F04 & \bm{36}(16) & \bm{19}(20) & \bm{19}(19) & \bm{17}(18) & \bm{18}(18) & \bm{18}(17) & \bm{16}(10) & \bm{14}(14) & \bm{12}(15) & \bm{12}(14) \\
            F05 & 65(40) & 69(40) & 74($\textendash$) & $\textendash$ & $\textendash$ & $\textendash$ & $\textendash$ & $\textendash$ & $\textendash$ & $\textendash$ \\
            F06 & \bm{39}(18) & \bm{28}(11) & \bm{28}(16) & \bm{31}(14) & \bm{27}(9) & \bm{27}(14) & \bm{31}(20) & \bm{35}(19) & \bm{39}(19) & \bm{47}(59) \\
            F07 & \bm{36}(27) & \bm{31}(12) & \bm{29}(14) & \bm{30}(13) & \bm{33}(13) & \bm{33}(13) & \bm{37}(17) & \bm{35}(26) & \bm{37}(19) & \bm{34}(17) \\
            F08 & \bm{39}(11) & \bm{34}(9) & \bm{26}(7) & \bm{25}(13) & \bm{25}(9) & \bm{24}(12) & \bm{28}(15) & \bm{27}(22) & \bm{45}(59) & 62($\textendash$) \\
            F09 & \bm{37}(12) & \bm{25}(6) & \bm{22}(13) & \bm{19}(14) & \bm{15}(8) & \bm{14}(10) & \bm{10}(7) & \bm{8}(3) & \bm{8}(4) & \bm{5}(5) \\
            F10 & \bm{45}(23) & \bm{49}(42) & \bm{39}(34) & \bm{42}(23) & 59(42) & \bm{46}(36) & 52(35) & 73($\textendash$) & 58($\textendash$) & $\textendash$ \\
            F11 & 65($\textendash$) & 56(78) & \bm{45}(58) & 71($\textendash$) & 68($\textendash$) & 81($\textendash$) & 76($\textendash$) & 66($\textendash$) & 94($\textendash$) & 50(74) \\
            F12 & \bm{37}(21) & \bm{26}(12) & \bm{24}(14) & \bm{23}(8) & \bm{18}(11) & \bm{19}(10) & \bm{15}(10) & \bm{19}(14) & \bm{19}(7) & \bm{18}(14) \\
            F13 & \bm{47}(7) & \bm{45}(13) & 50(9) & 51(15) & \bm{48}(15) & \bm{49}(10) & 54(22) & 85(33) & 91($\textendash$) & $\textendash$ \\
            F14 & \bm{39}(16) & \bm{29}(15) & \bm{30}(11) & \bm{29}(6) & \bm{32}(13) & \bm{33}(10) & \bm{31}(16) & \bm{35}(27) & \bm{42}(21) & 50(52) \\
            F15 & \bm{24}(24) & \bm{12}(10) & \bm{13}(13) & \bm{13}(9) & \bm{13}(15) & \bm{13}(11) & \bm{12}(10) & \bm{17}(15) & \bm{15}(14) & \bm{14}(10) \\
            F16 & 50(62) & 67($\textendash$) & 94($\textendash$) & 55($\textendash$) & $\textendash$ & $\textendash$ & $\textendash$ & $\textendash$ & $\textendash$ & $\textendash$ \\
            F17 & \bm{26}(22) & \bm{18}(20) & \bm{15}(21) & \bm{15}(23) & \bm{13}(23) & \bm{14}(11) & \bm{12}(14) & \bm{10}(12) & \bm{14}(9) & \bm{11}(8) \\
            F18 & \bm{26}(29) & \bm{21}(23) & \bm{16}(20) & \bm{15}(28) & \bm{18}(13) & \bm{13}(17) & \bm{13}(13) & \bm{15}(14) & \bm{14}(12) & \bm{12}(16) \\
            F19 & \bm{15}(14) & \bm{10}(10) & \bm{10}(9) & \bm{9}(7) & \bm{10}(12) & \bm{8}(6) & \bm{9}(8) & \bm{9}(9) & \bm{8}(8) & \bm{6}(10) \\
            F20 & \bm{43}(11) & \bm{31}(10) & \bm{30}(11) & \bm{28}(9) & \bm{29}(11) & \bm{29}(9) & \bm{35}(14) & \bm{45}(35) & 63($\textendash$) & $\textendash$ \\
            F21 & \bm{48}($\textendash$) & \bm{38}($\textendash$) & 74($\textendash$) & 82($\textendash$) & 94($\textendash$) & 56($\textendash$) & 85($\textendash$) & $\textendash$ & 74($\textendash$) & $\textendash$ \\
            F22 & 69($\textendash$) & 63($\textendash$) & \bm{47}(53) & 52(72) & 56(54) & 54(46) & \bm{42}(46) & 65(48) & 58(32) & 69(24) \\
            F23 & \bm{38}(42) & 70($\textendash$) & 63($\textendash$) & 61($\textendash$) & 60($\textendash$) & \bm{36}(61) & \bm{44}($\textendash$) & \bm{35}(51) & 51($\textendash$) & \bm{41}($\textendash$) \\
            F24 & \bm{31}(36) & \bm{16}(25) & \bm{12}(16) & \bm{14}(12) & \bm{10}(7) & \bm{11}(15) & \bm{12}(15) & \bm{10}(12) & \bm{9}(13) & \bm{5}(8) \\
            \hline\hline
            \multicolumn{1}{c}{Median} &
            \multicolumn{1}{c}{38.5} &
            \multicolumn{1}{c}{30.8} &
            \multicolumn{1}{c}{29.0} &
            \multicolumn{1}{c}{29.0} &
            \multicolumn{1}{c}{28.0} &
            \multicolumn{1}{c}{27.8} &
            \multicolumn{1}{c}{30.8} &
            \multicolumn{1}{c}{34.5} &
            \multicolumn{1}{c}{38.5} &
            \multicolumn{1}{c}{34.0}\\
            \hline \hline
        \end{tabular}
    \end{center}
    \vspace{1em}
    \raggedright{
        We calculated the number of iterations required for PolyBO to reach the simple regret reached by vanilla BO at 50 iterations.
        Each value is the median number of iterations at which PolyBO reached the reference simple regret, and the values in parentheses indicate the interquartile range.
        Values $<50$ are in bold, and values $\geq50$ are in plain font.
        The maximum number of iterations was set to 100; \textendash{} indicates that the value could not be calculated because the reference simple regret was not reached within 100 iterations.
        $D=20$, $p=4$.
    }
\end{table}

\clearpage
\begin{table}[b]
    \begin{center}
        \caption{Sensitivity analysis of the maximum polynomial degree $p$ in PolyBO}
        \label{tb:sensitivity_p}
        \vspace{1em}
        \begin{tabular}{l *{9}{D{(}{\,(}{3}}}
            \hline
            \multicolumn{1}{l}{ID} &
            \multicolumn{1}{c}{$p=1$} &
            \multicolumn{1}{c}{$p=2$} &
            \multicolumn{1}{c}{$p=3$} &
            \multicolumn{1}{c}{$p=4$} &
            \multicolumn{1}{c}{$p=5$} &
            \multicolumn{1}{c}{$p=6$} &
            \multicolumn{1}{c}{$p=7$} &
            \multicolumn{1}{c}{$p=8$} \\
            \hline
            F01 & 72(9) & \bm{48}(6) & $\textendash$ & \bm{49}(12) & \bm{48}(6) & 63(36) & $\textendash$ & $\textendash$ \\
            F02 & 55(22) & \bm{43}(19) & 67(39) & \bm{39}(18) & \bm{43}(19) & \bm{39}(19) & \bm{48}(33) & 50(28) \\
            F03 & \bm{38}(8) & \bm{30}(10) & \bm{30}(16) & \bm{23}(9) & \bm{30}(10) & \bm{21}(18) & \bm{22}(14) & \bm{27}(15) \\
            F04 & \bm{48}(46) & \bm{26}(19) & \bm{36}(36) & \bm{19}(19) & \bm{26}(19) & \bm{19}(22) & \bm{24}(23) & \bm{19}(18) \\
            F05 & 55(7) & 71(25) & 73($\textendash$) & 74($\textendash$) & 71(25) & 78($\textendash$) & 75($\textendash$) & 79($\textendash$) \\
            F06 & \bm{43}(5) & \bm{38}(15) & \bm{32}(14) & \bm{28}(16) & \bm{38}(15) & \bm{30}(16) & \bm{30}(21) & \bm{31}(16) \\
            F07 & \bm{45}(11) & \bm{37}(18) & \bm{32}(25) & \bm{29}(14) & \bm{37}(18) & \bm{32}(18) & \bm{32}(13) & \bm{32}(16) \\
            F08 & 52(17) & \bm{32}(9) & \bm{45}(19) & \bm{26}(7) & \bm{32}(9) & \bm{28}(11) & \bm{41}(36) & \bm{29}(20) \\
            F09 & $\textendash$ & \bm{26}(7) & \bm{38}(18) & \bm{22}(13) & \bm{26}(7) & \bm{22}(11) & \bm{26}(16) & \bm{17}(13) \\
            F10 & 83($\textendash$) & 53(39) & 100($\textendash$) & \bm{39}(34) & 53(39) & \bm{38}(23) & 57($\textendash$) & \bm{44}(46) \\
            F11 & 84($\textendash$) & 73($\textendash$) & 74($\textendash$) & \bm{45}(58) & 73($\textendash$) & 52($\textendash$) & 84($\textendash$) & $\textendash$ \\
            F12 & \bm{38}(12) & \bm{28}(12) & \bm{24}(13) & \bm{24}(14) & \bm{28}(12) & \bm{24}(12) & \bm{23}(13) & \bm{24}(10) \\
            F13 & 70(8) & \bm{47}(8) & 72(26) & 50(9) & \bm{47}(8) & 52(14) & $\textendash$ & 82($\textendash$) \\
            F14 & 53(21) & \bm{34}(15) & \bm{39}(26) & \bm{30}(11) & \bm{34}(15) & \bm{31}(13) & \bm{39}(21) & \bm{36}(25) \\
            F15 & \bm{32}(13) & \bm{13}(15) & \bm{17}(12) & \bm{13}(13) & \bm{13}(15) & \bm{10}(12) & \bm{11}(11) & \bm{9}(5) \\
            F16 & 83($\textendash$) & $\textendash$ & $\textendash$ & 94($\textendash$) & 91($\textendash$) & \bm{47}(59) & 93($\textendash$) & 63(61) \\
            F17 & \bm{23}(19) & \bm{21}(17) & \bm{14}(17) & \bm{15}(21) & \bm{17}(14) & \bm{19}(16) & \bm{10}(16) & \bm{14}(15) \\
            F18 & \bm{23}(16) & \bm{21}(22) & \bm{10}(18) & \bm{16}(20) & \bm{13}(14) & \bm{15}(17) & \bm{11}(20) & \bm{20}(31) \\
            F19 & \bm{42}(9) & \bm{28}(22) & \bm{16}(13) & \bm{10}(9) & \bm{10}(8) & \bm{10}(11) & \bm{9}(7) & \bm{8}(7) \\
            F20 & 61(17) & \bm{35}(14) & \bm{41}(16) & \bm{30}(11) & \bm{39}(15) & \bm{28}(12) & \bm{47}(31) & \bm{37}(18) \\
            F21 & \bm{41}($\textendash$) & \bm{32}(18) & 53($\textendash$) & 74($\textendash$) & \bm{40}($\textendash$) & \bm{44}($\textendash$) & \bm{41}($\textendash$) & \bm{44}(69) \\
            F22 & 52(58) & 51(36) & 77($\textendash$) & \bm{47}(53) & \bm{48}(48) & 51(75) & \bm{31}(42) & \bm{38}(41) \\
            F23 & 70($\textendash$) & \bm{35}(84) & 52($\textendash$) & 63($\textendash$) & \bm{38}(86) & 58($\textendash$) & \bm{46}(76) & \bm{29}(54) \\
            F24 & \bm{37}(30) & \bm{24}(26) & \bm{40}(27) & \bm{12}(16) & \bm{13}(11) & \bm{12}(16) & \bm{14}(14) & \bm{10}(10) \\
            \hline\hline
            \multicolumn{1}{c}{Median} &
            \multicolumn{1}{c}{51.5} &
            \multicolumn{1}{c}{33.5} &
            \multicolumn{1}{c}{39.3} &
            \multicolumn{1}{c}{29.0} &
            \multicolumn{1}{c}{37.3} &
            \multicolumn{1}{c}{30.3} &
            \multicolumn{1}{c}{31.3} &
            \multicolumn{1}{c}{30.0} \\
            \hline \hline
        \end{tabular}
    \end{center}
    \vspace{1em}
    \raggedright{
        We calculated the number of iterations required for PolyBO to reach the simple regret reached by vanilla BO at 50 iterations.
        Each value is the median number of iterations at which PolyBO reached the reference simple regret, and the values in parentheses indicate the interquartile range.
        Values $<50$ are in bold and values $\geq50$ are in plain font.
        The maximum number of iterations was set to 100; \textendash{} indicates that the value could not be calculated because the reference simple regret was not reached within 100 iterations.
        $D=20$, $m'=10$.
    }
\end{table}

\clearpage

\clearpage

\clearpage

\clearpage

\setcounter{figure}{0}
\setcounter{table}{0}
\setcounter{algorithm}{0}
\renewcommand{\theHfigure}{supplementary.\arabic{figure}}
\renewcommand{\theHtable}{supplementary.\arabic{table}}
\providecommand{\theHalgorithm}{}
\renewcommand{\theHalgorithm}{supplementary.\arabic{algorithm}}
\floatname{algorithm}{Supplementary Algorithm}

\makeatletter
\renewcommand{\fnum@figure}{\textbf{Supplementary Figure\nobreakspace\thefigure.}}
\renewcommand{\fnum@table}{\textbf{Supplementary Table\nobreakspace\thetable.}}
\makeatother

\newlength{\mainfooterposition}
\setlength{\mainfooterposition}{\dimexpr\topmargin+\footskip\relax}
\setlength{\topmargin}{-3cm}
\setlength{\footskip}{\dimexpr\mainfooterposition-\topmargin\relax}

\clearpage
\begin{figure}[h!]
    \centering
    \includegraphics[scale=0.8]{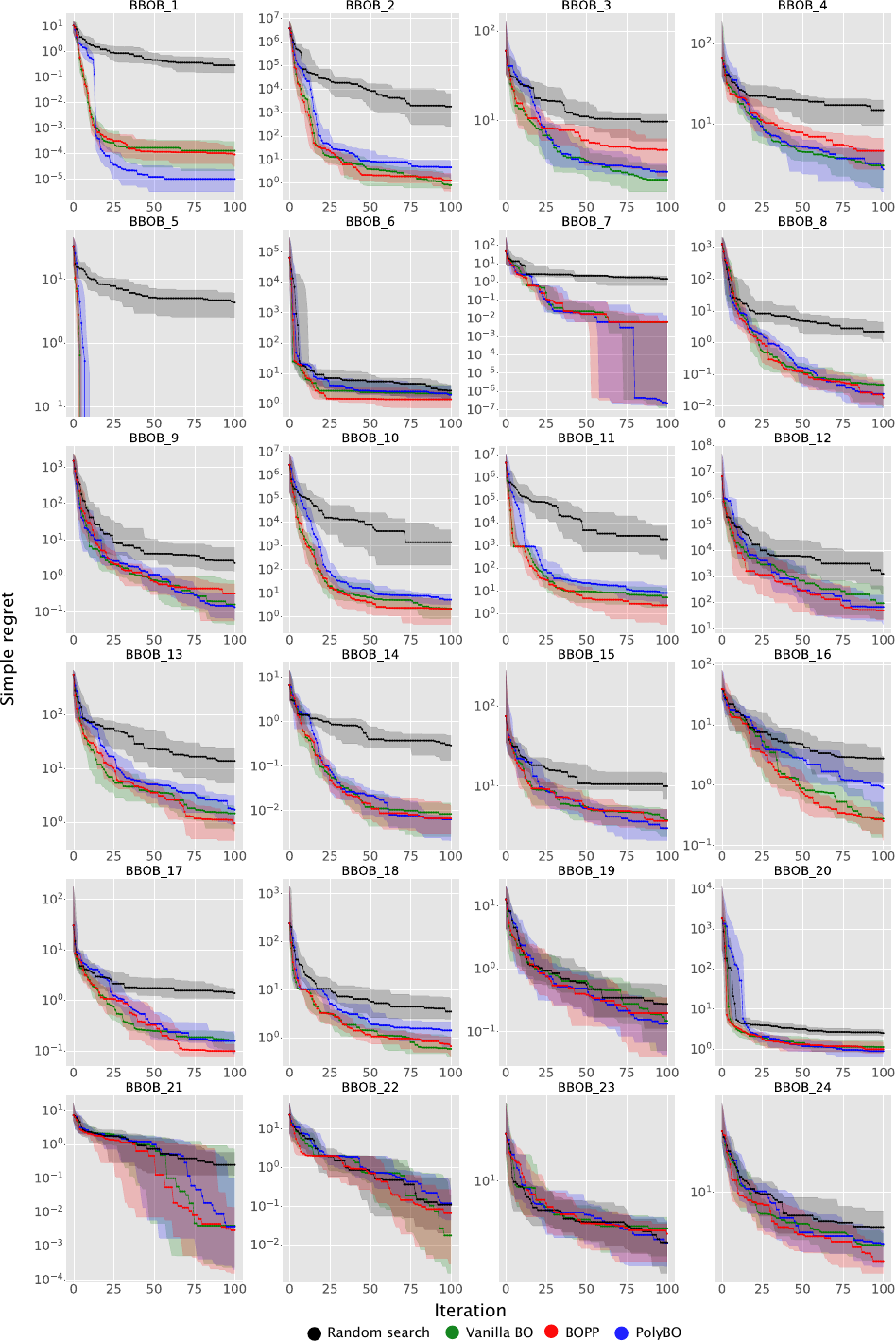}
    \caption{Optimization processes for 24 benchmark functions at $D=2$}
    \label{fig:supp_benchmark_d2}
    \raggedright{
        The median (lines) and interquartile range (shading) of 30 runs of optimization from different initial conditions are shown for each method.
        Problem setting: $m' = 10$, $p = 4$.
    }
\end{figure}

\clearpage
\begin{figure}[h!]
    \centering
    \includegraphics[scale=0.8]{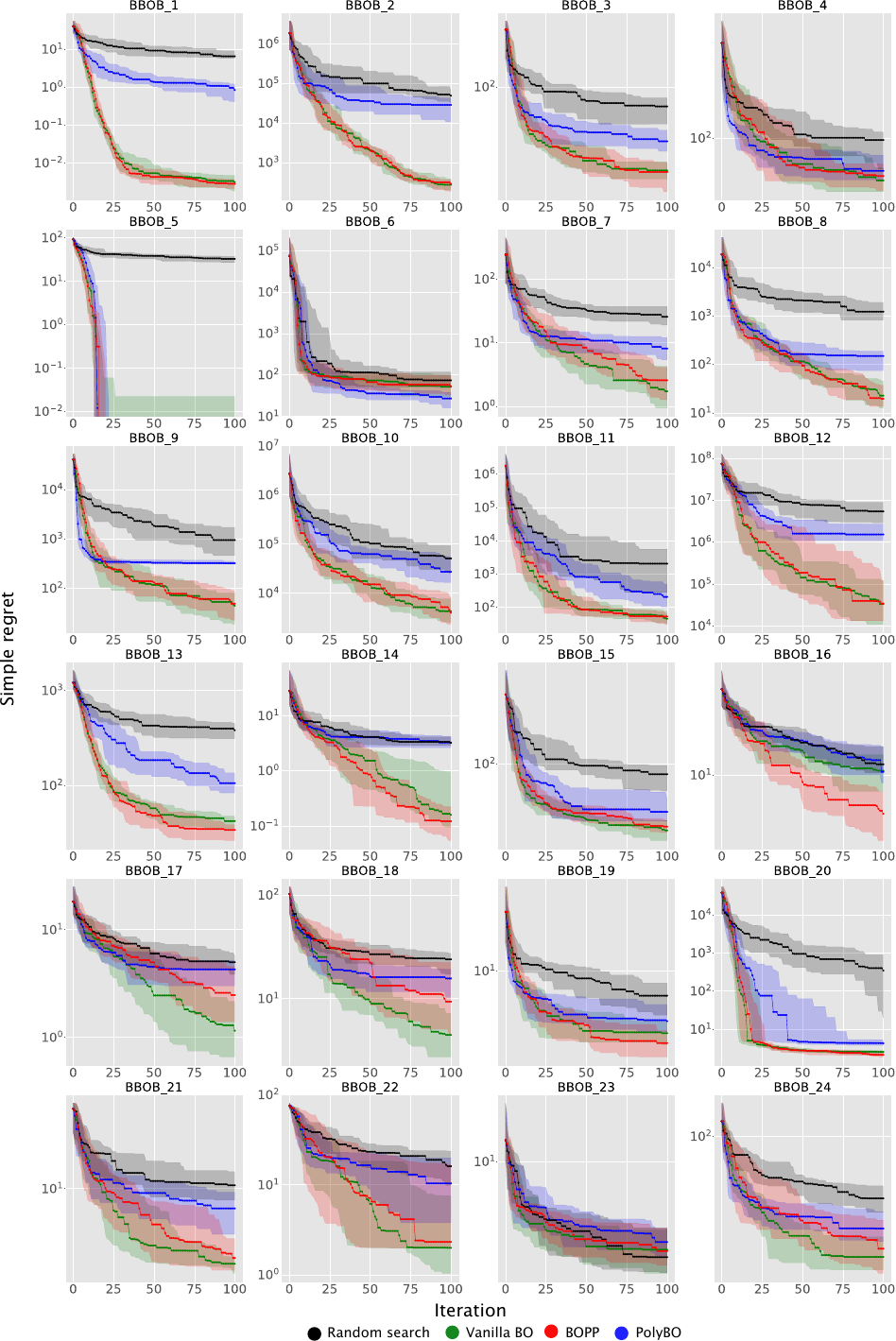}
    \caption{Optimization processes for 24 benchmark functions at $D=5$}
    \label{fig:supp_benchmark_d5}
    \raggedright{
        The median (lines) and interquartile range (shading) of 30 runs of optimization from different initial conditions are shown for each method.
        Problem setting: $m' = 10$, $p = 4$.
    }
\end{figure}

\clearpage
\begin{figure}[h!]
    \centering
    \includegraphics[scale=0.8]{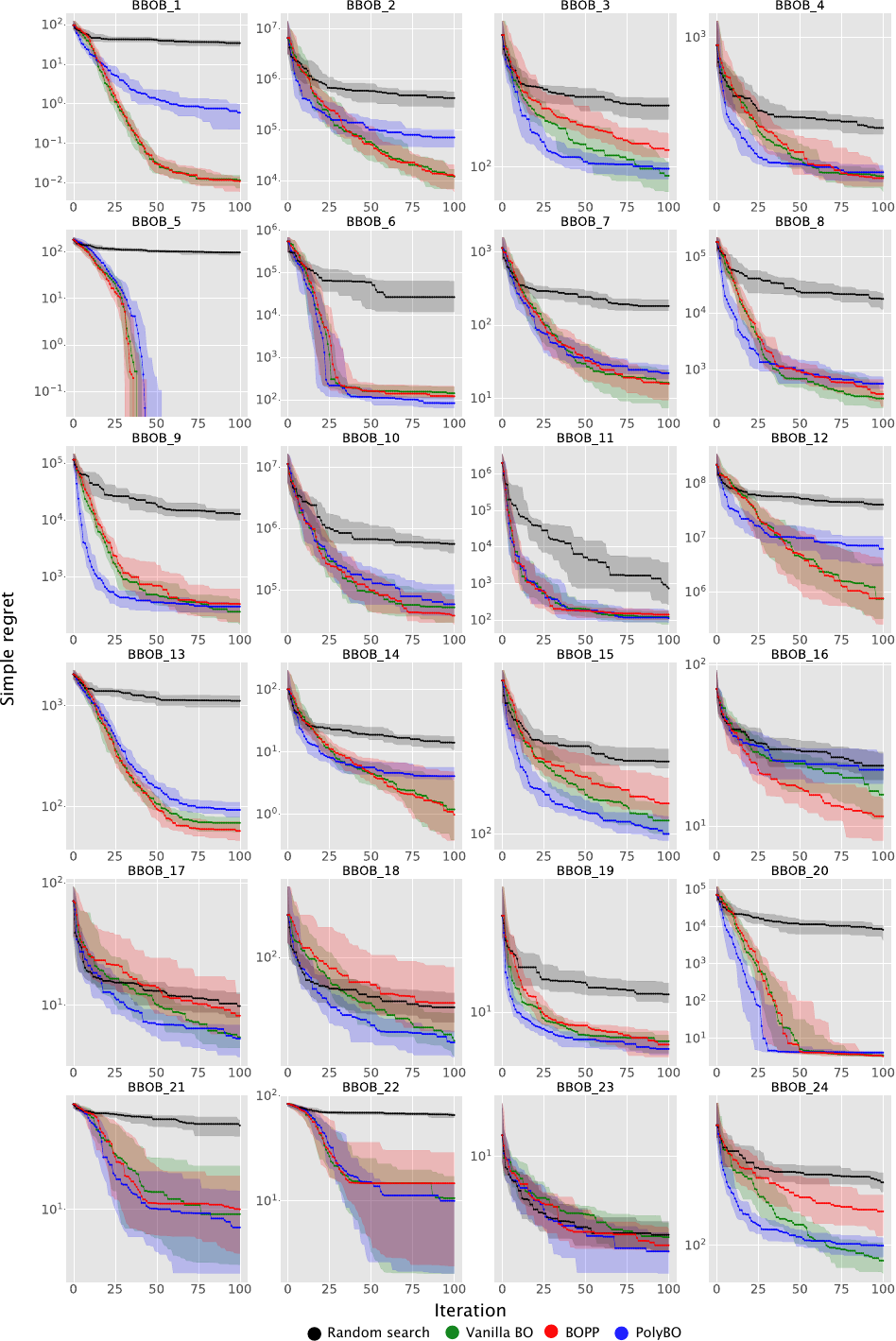}
    \caption{Optimization processes for 24 benchmark functions at $D=10$}
    \label{fig:supp_benchmark_d10}
    \raggedright{
        The median (lines) and interquartile range (shading) of 30 runs of optimization from different initial conditions are shown for each method.
        Problem setting: $m' = 10$, $p = 4$.
    }
\end{figure}

\clearpage
\begin{figure}[h!]
    \centering
    \includegraphics[scale=0.8]{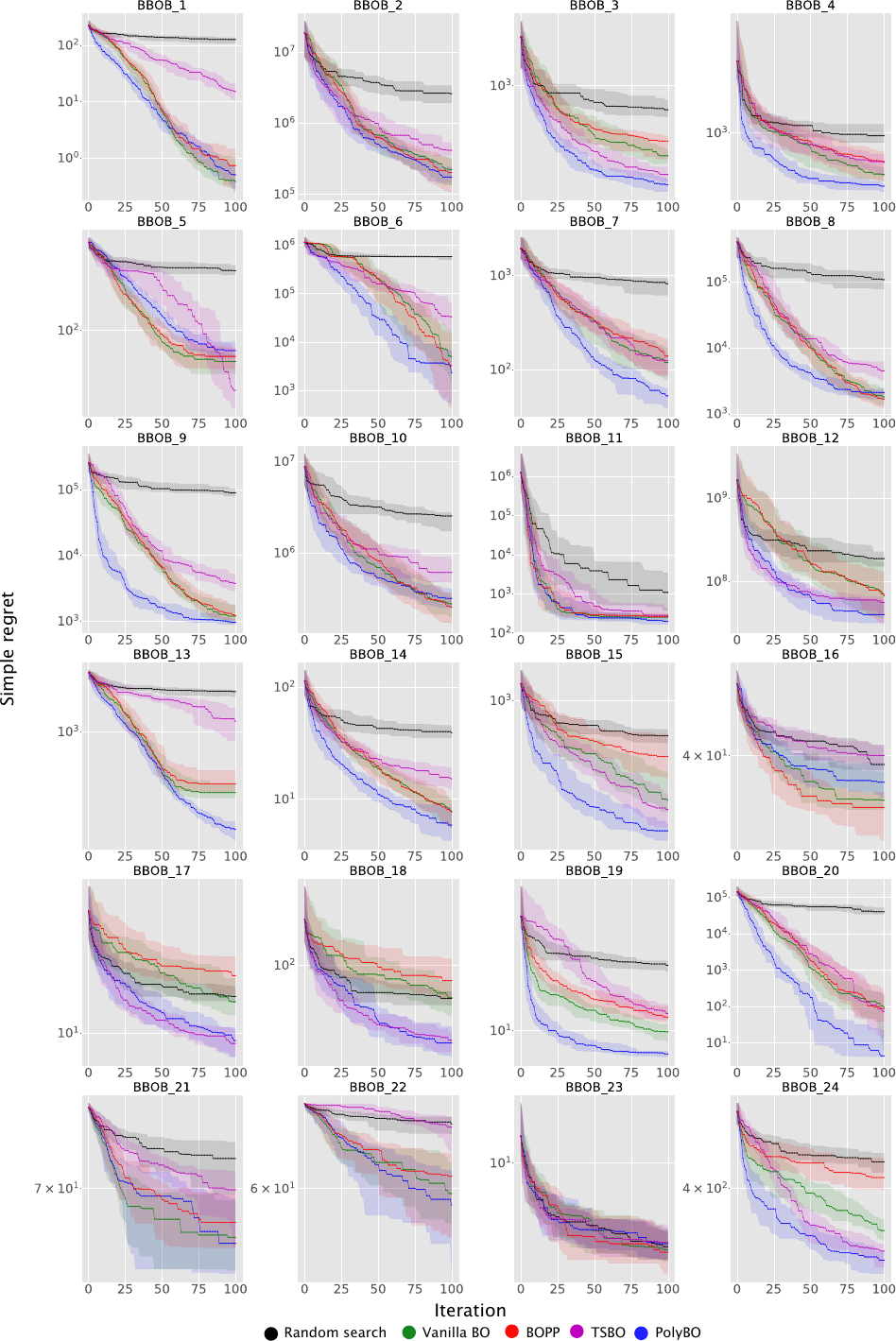}
    \caption{Optimization processes for 24 benchmark functions at $D=20$}
    \label{fig:supp_benchmark_d20}
    \raggedright{
        The median (lines) and interquartile range (shading) of 30 runs of optimization from different initial conditions are shown for each method.
        Problem setting: $m' = 10$, $p = 4$.
    }
\end{figure}

\clearpage
\begin{figure}[h]
    \centering
    \includegraphics[width=17cm]{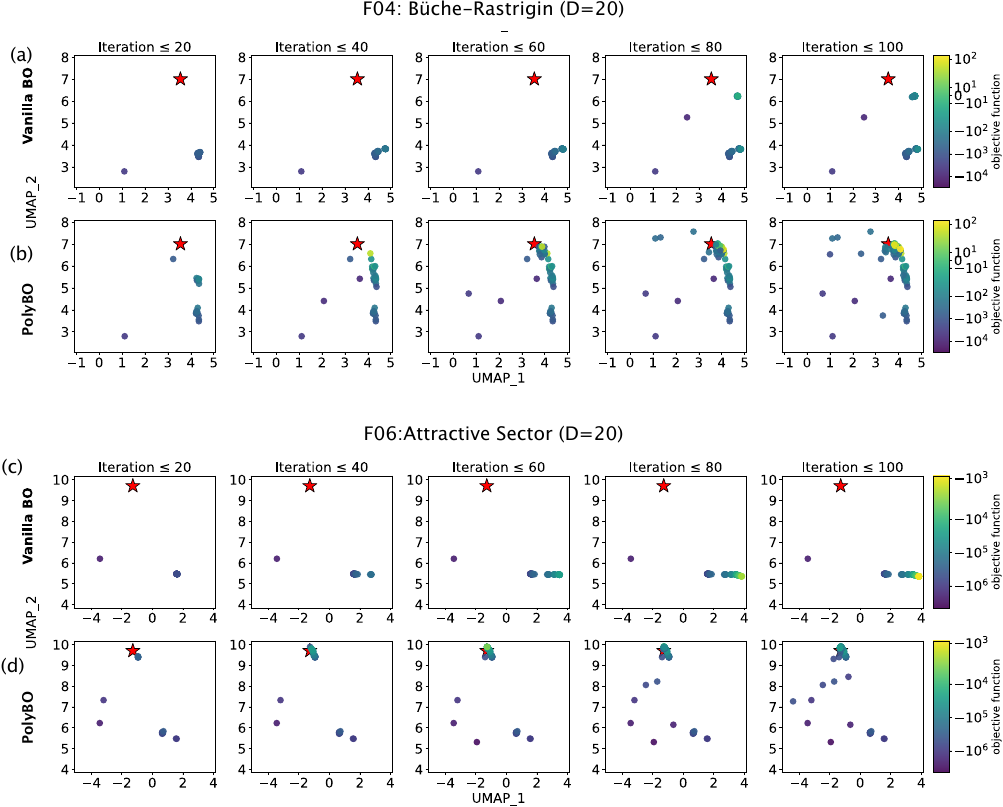}
    \caption{UMAP visualization of optimization processes for F04 and F06 at $D=20$}
    \label{fig:supp_umap_d20}
    \raggedright{
        Circles indicate all sampled points used in the optimization, including the initial experimental points and the candidate points $\mathbf{x}^{(t)}$ selected by each method at subsequent iterations.
        Color indicates the value of the objective function.
        A symlog scale was used for the color bar, and the threshold for the linear region was set to $10^{-3}$ of the maximum value of the absolute plotted objective function.
        Red stars indicate the locations of the global optima.
        $m' = 10$, $p = 4$.
    }
\end{figure}

\clearpage
\begin{figure}[h]
    \centering
    \includegraphics[width=17cm]{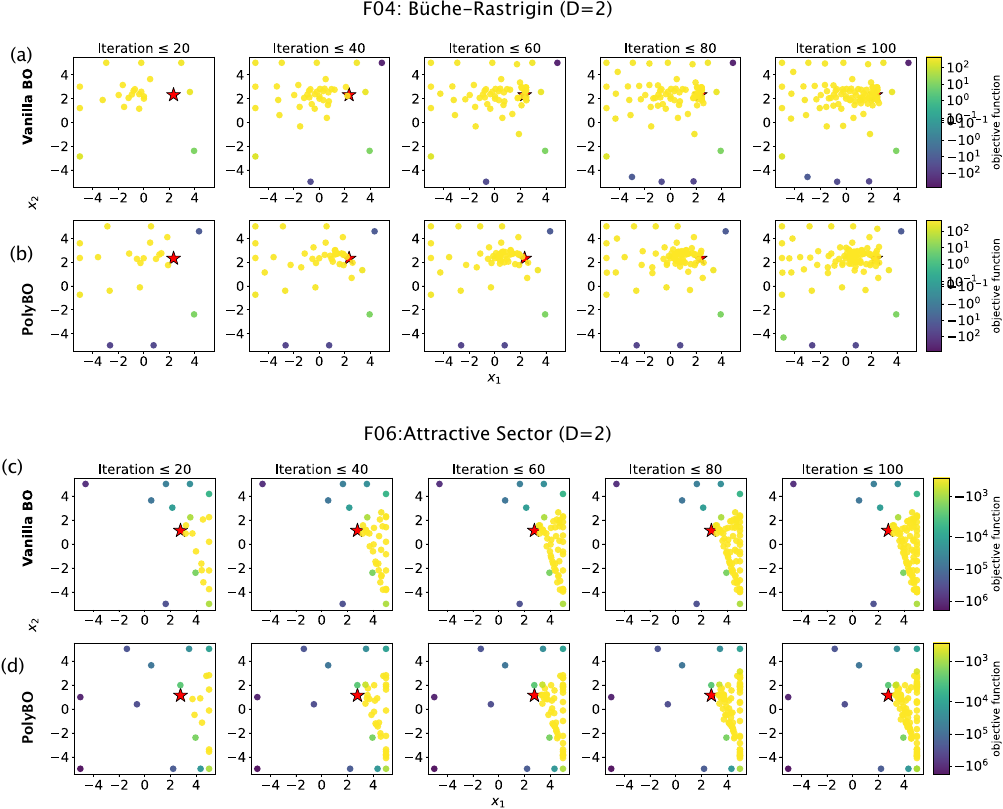}
    \caption{t-SNE visualization of optimization processes for F04 and F06 at $D=2$}
    \label{fig:supp_tsne_d2}
    \raggedright{
        Circles indicate all sampled points used in the optimization, including the initial experimental points and the candidate points $\mathbf{x}^{(t)}$ selected by each method at subsequent iterations.
        Color indicates the value of the objective function.
        A symlog scale was used for the color bar, and the threshold for the linear region was set to $10^{-3}$ of the maximum absolute plotted objective function value.
        Red stars indicate the locations of the global optima.
        $m' = 10$, $p = 4$.
    }
\end{figure}

\clearpage
\begin{figure}[h]
    \centering
    \includegraphics[width=17cm]{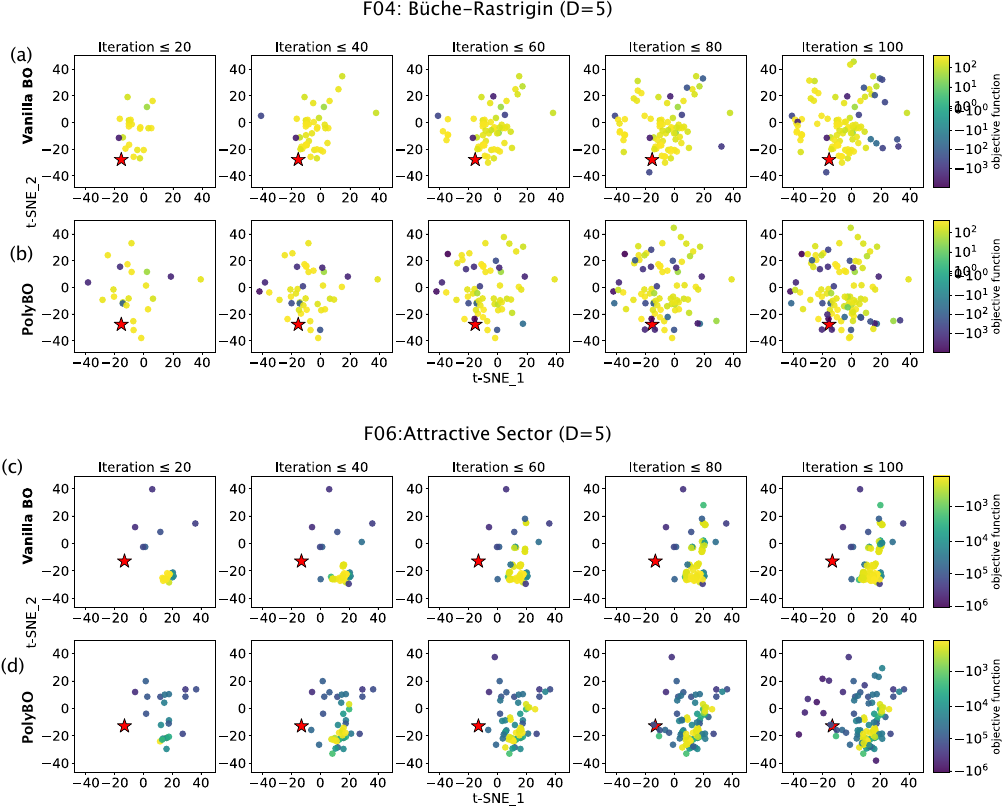}
    \caption{t-SNE visualization of optimization processes for F04 and F06 at $D=5$}
    \label{fig:supp_tsne_d5}
    \raggedright{
        Circles indicate all sampled points used in the optimization, including the initial experimental points and the candidate points $\mathbf{x}^{(t)}$ selected by each method at subsequent iterations.
        Color indicates the value of the objective function.
        A symlog scale was used for the color bar, and the threshold for the linear region was set to $10^{-3}$ of the maximum absolute plotted objective function value.
        Red stars indicate the locations of the global optima.
        $m' = 10$, $p = 4$.
    }
\end{figure}

\clearpage
\begin{figure}[h]
    \centering
    \includegraphics[width=10cm]{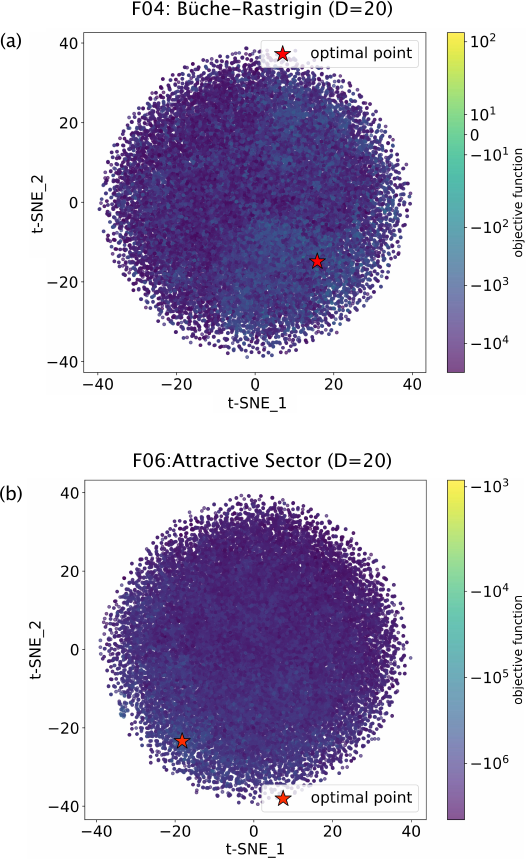}
    \caption{Entire search space after dimensionality reduction using t-SNE}
    \label{fig:supp_tsne_landscape}
    \raggedright{
        A symlog scale was used for the color bar, and the threshold for the linear region was set to $10^{-3}$ of the value of the maximum absolute plotted objective function.
        Color indicates the value of the objective function.
        Red stars indicate the locations of the global optima.
        $m' = 10$, $p = 4$.
    }
\end{figure}

\clearpage
\begin{figure}[h]
    \centering
    \includegraphics[width=16cm]{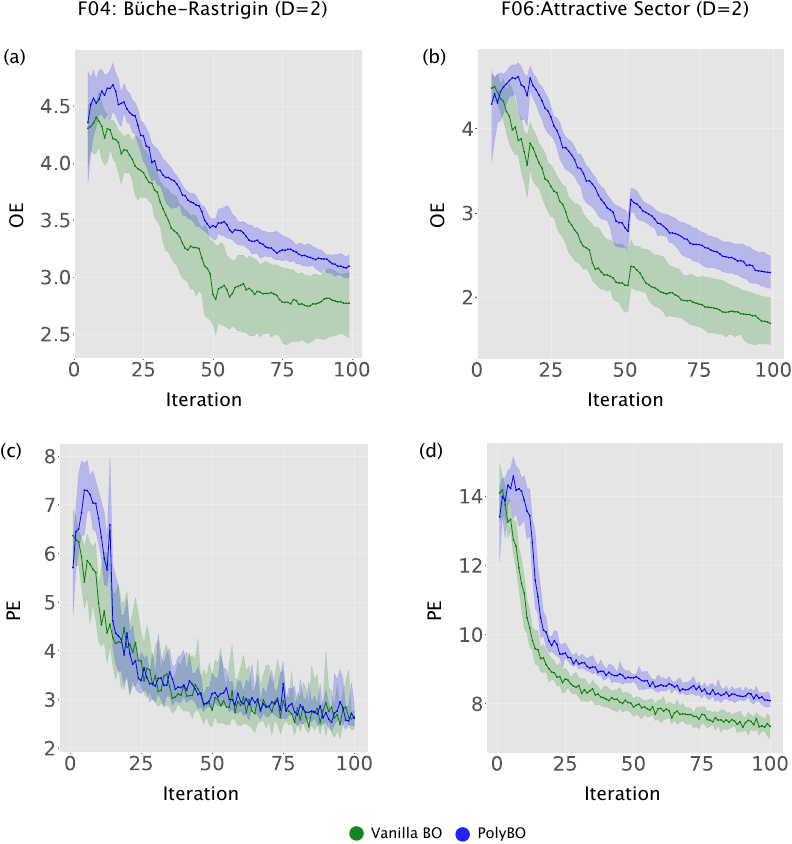}
    \caption{Quantification of optimization behavior for F04 and F06 at $D=2$ using OE and PE}
    \label{fig:supp_entropy_d2}
    \raggedright{
        The median (lines) and interquartile range (shading) of 30 runs of optimization from different initial conditions are shown for each method.
        $m' = 10$, $p = 4$.
    }
\end{figure}

\clearpage
\begin{figure}[h]
    \centering
    \includegraphics[width=16cm]{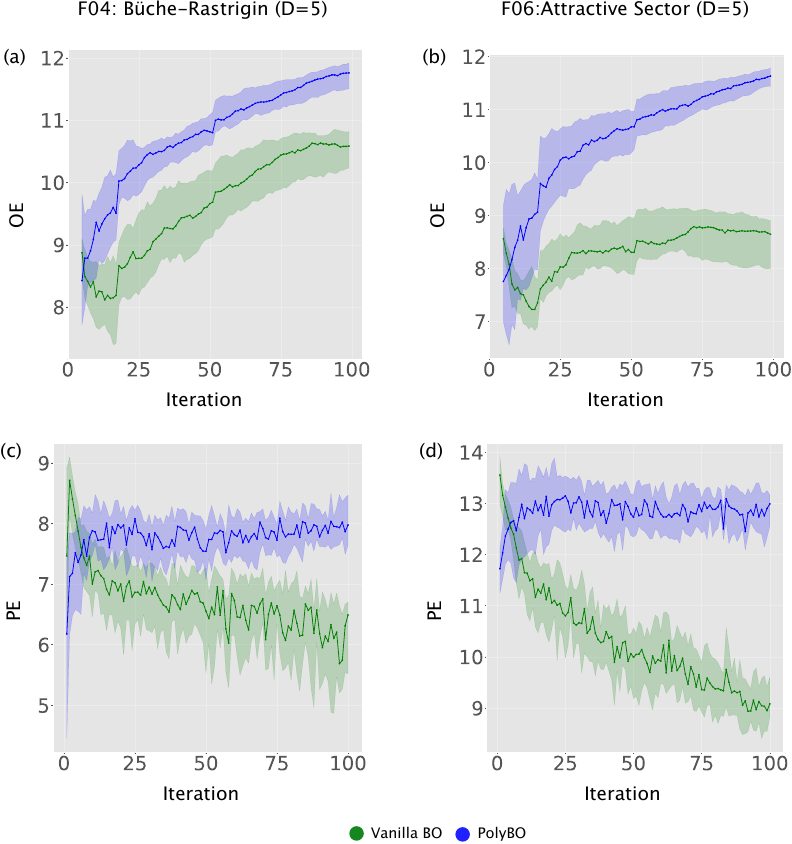}
    \caption{Quantification of optimization behavior for F04 and F06 at $D=5$ using OE and PE}
    \label{fig:supp_entropy_d5}
    \raggedright{
        The median (lines) and interquartile range (shading) of 30 runs of optimization from different initial conditions are shown for each method.
        $m' = 10$, $p = 4$.
    }
\end{figure}

\clearpage
\begin{algorithm}
    \caption[BO using pseudo-experimental data under discrete constraints]{PolyBO algorithm for a discrete constrained search space}
    \label{algo:pseudo_experiment_discrete_bo}
    \begin{algorithmic}[1]
        \REQUIRE Discrete grid interval $\delta$, bounds $[l_i, u_i]\ (i=1,\dots,d)$ for each variable, equality constraint $\sum_{i=1}^{d} x_i = 1$
        \STATE Uniformly sample $k$ explainable variables from the discrete grid set $\mathcal{G}_\delta$ satisfying $\sum_{i=1}^{d} x_i = 1$ and $l_i \le x_i \le u_i$, and construct the initial experimental dataset $\mathcal{D}_{0:0} = \{(\mathbf{x}^{(0)}_i, y^{(0)}_i)\}_{i=1}^k$.
        \FOR {$t = 1, 2, \dots$}
        \STATE Train $p^{\mathrm{th}}$ degree polynomial regression model $f_p(\mathbf{x})$ using dataset $\mathcal{D}_{0:t-1}$.
        \STATE Uniformly sample $m'$ explainable variables $\{\mathbf{x'}^{(t)}_j\}_{j=1}^{m'}$ from the discrete grid set $\mathcal{G}_\delta$.
        \STATE Compute $y'^{(t)}_j = f_p(\mathbf{x'}^{(t)}_j)$ for $j = 1, \dots, m'$, and construct pseudo-experimental data $\mathcal{D}_t' = \{(\mathbf{x'}^{(t)}_j, y'^{(t)}_j)\}_{j=1}^{m'}$.
        \STATE Update the Gaussian process regression model with $\mathcal{D}_{0:t-1} \cup \mathcal{D}_t'$.
        \STATE Compute the acquisition function $\alpha(\mathbf{x})$ using the updated Gaussian process regression model.
        \STATE Optimize $\alpha(\mathbf{x})$ in a continuous space under the constraint $\sum_{i=1}^{d} x_i = 1$ using the SLSQP method from multiple initial conditions.
        \STATE For each continuous optimum, enumerate adjacent lower and upper discrete grid points in each dimension, and extract only combinations satisfying $\sum_{i=1}^{d} x_i = 1$ as discrete candidates.
        \STATE Re-evaluate $\alpha(\mathbf{x})$ for all discrete candidates and select the unobserved candidate with the maximum value as $\mathbf{x}^{(t)}$.
        \STATE Conduct an experiment at $\mathbf{x}^{(t)}$ to obtain $y^{(t)}$.
        \STATE Update the dataset $\mathcal{D}_{0:t} = \mathcal{D}_{0:t-1} \cup \{(\mathbf{x}^{(t)}, y^{(t)})\}$.
        \ENDFOR
    \end{algorithmic}
\end{algorithm}

\begin{figure}[h]
    \centering
    \includegraphics[width=17cm]{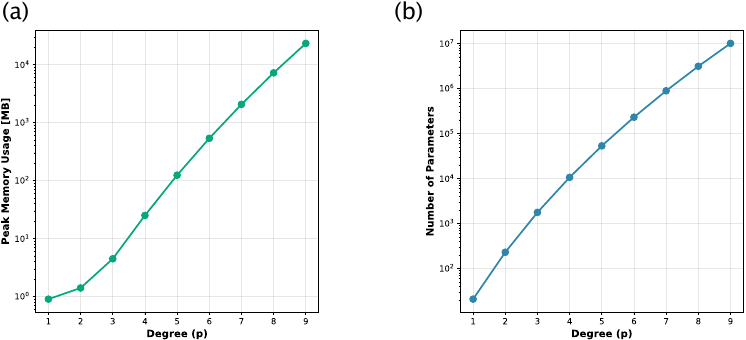}
    \caption{Peak memory usage and number of parameters at different maximum polynomial degrees}
    \label{fig:supp_memory}
    \raggedright{
        Peak memory usage (a) was measured to evaluate memory complexity.
    }
\end{figure}

\end{document}